\documentclass[10pt,twocolumn,letterpaper]{article}

\usepackage[pagenumbers]{cvpr} 
\usepackage[accsupp]{axessibility}  

\usepackage{array}
\newcommand{\PreserveBackslash}[1]{\let\temp=\\#1\let\\=\temp}
\newcolumntype{C}[1]{>{\PreserveBackslash\centering}p{#1}}
\newcolumntype{R}[1]{>{\PreserveBackslash\raggedleft}p{#1}}
\newcolumntype{L}[1]{>{\PreserveBackslash\raggedright}p{#1}}
\usepackage{graphicx}
\usepackage{amsmath}
\usepackage{amssymb}
\usepackage{booktabs}
\usepackage{multirow}
\usepackage{multicol}
\usepackage{arydshln}
\usepackage[dvipsnames]{xcolor}
\usepackage{bbm}
\usepackage{float}
\usepackage{pifont}
\newcommand{\cmark}{\ding{51}}%
\newcommand{\xmark}{\ding{55}}%
\definecolor{grey}{RGB}{128,128,128}
\def\greyColor{\textcolor{grey}}
\def\ft{\textcolor{grey}}

\usepackage{amsthm}   

\newtheorem{thm}{Theorem}

\newtheorem{ass}{Assumption}

\def \E {\mathrm{E}}
\def \Lmath {\mathcal{L}}
\def \D {\mathcal{D}}

\def \F {\mathcal{F}}
\def \R {\mathbb{R}}

\usepackage[pagebackref,breaklinks,colorlinks]{hyperref}

\usepackage[capitalize]{cleveref}
\crefname{section}{Sec.}{Secs.}
\Crefname{section}{Section}{Sections}
\Crefname{table}{Table}{Tables}
\crefname{table}{Tab.}{Tabs.}

\usepackage{color}

\newcommand{\tablestyle}[2]{\setlength{\tabcolsep}{#1}\renewcommand{\arraystretch}{#2}\centering\small}
\newcommand{\tabincell}[2]{\begin{tabular}{@{}#1@{}}#2\end{tabular}}  
\newlength\savewidth\newcommand\shline{\noalign{\global\savewidth\arrayrulewidth
  \global\arrayrulewidth 1pt}\hline\noalign{\global\arrayrulewidth\savewidth}}
\def\cvprPaperID{1579} 
\def\confName{CVPR}
\def\confYear{2022}

\begin{document}


\title{Learning from Untrimmed Videos:\\ Self-Supervised Video Representation Learning with Hierarchical Consistency}
\author{Zhiwu Qing$^{1}$ \quad Shiwei Zhang$^{2*}$ \quad Ziyuan Huang$^{3}$ \quad Yi Xu$^{4}$ \quad Xiang Wang$^{1}$ \\ \quad Mingqian Tang$^2$ 
\quad  Changxin Gao$^{1*}$  \quad Rong Jin$^2$   \quad Nong Sang$^1$
\\
$^1$Key Laboratory of Image Processing and Intelligent Control \\ School of Artificial Intelligence and Automation, Huazhong University of Science and Technology\\
$^2$Alibaba Group \quad $^3$ARC, National University of Singapore \quad $^4$Dalian Unversity of Technology\\

{\tt\small \{qzw, wxiang, cgao, nsang\}@hust.edu.cn}\\
{\tt\small \{zhangjin.zsw, mingqian.tmq, jinrong.jr\}@alibaba-inc.com} \\
{\tt\small ziyuan.huang@u.nus.edu} \quad {\tt\small yxu@dlut.edu.cn} \\

\vspace{-0.6cm}
}
\maketitle
\let\thefootnote\relax\footnotetext{$^*$Corresponding authors.}
\let\thefootnote\relax\footnotetext{Project page: \url{https://hico-cvpr2022.github.io/}.}
%

\begin{abstract}
\vspace{-0.2cm}

Natural videos provide rich visual contents for self-supervised learning. 
%
%
Yet most existing approaches for learning spatio-temporal representations rely on manually trimmed videos, leading to limited diversity in visual patterns and limited performance gain. 
In this work, we aim to learn representations by leveraging more abundant information in untrimmed videos.
%
%
To this end, we propose to learn a hierarchy of consistencies in videos, \textit{i.e.,} visual consistency and topical consistency, corresponding respectively to clip pairs that tend to be visually similar when separated by a short time span and share similar topics when separated by a long time span.
Specifically, a hierarchical consistency learning framework \textbf{HiCo} is presented, where the visually consistent pairs are encouraged to have the same representation through contrastive learning, while the topically consistent pairs are coupled through a topical classifier that distinguishes whether they are topic-related.
%
%
Further, we impose a gradual sampling algorithm for proposed hierarchical consistency learning, and demonstrate its theoretical superiority.
Empirically, we show that not only HiCo can generate stronger representations on untrimmed videos, it also improves the representation quality when applied to trimmed videos.
This is in contrast to standard contrastive learning that fails to learn appropriate representations from untrimmed videos.
%

\vspace{-5mm}
\end{abstract}
\section{Introduction}
\label{sec:intro}

\begin{figure}
    \centering
    \includegraphics[width=1.0\linewidth]{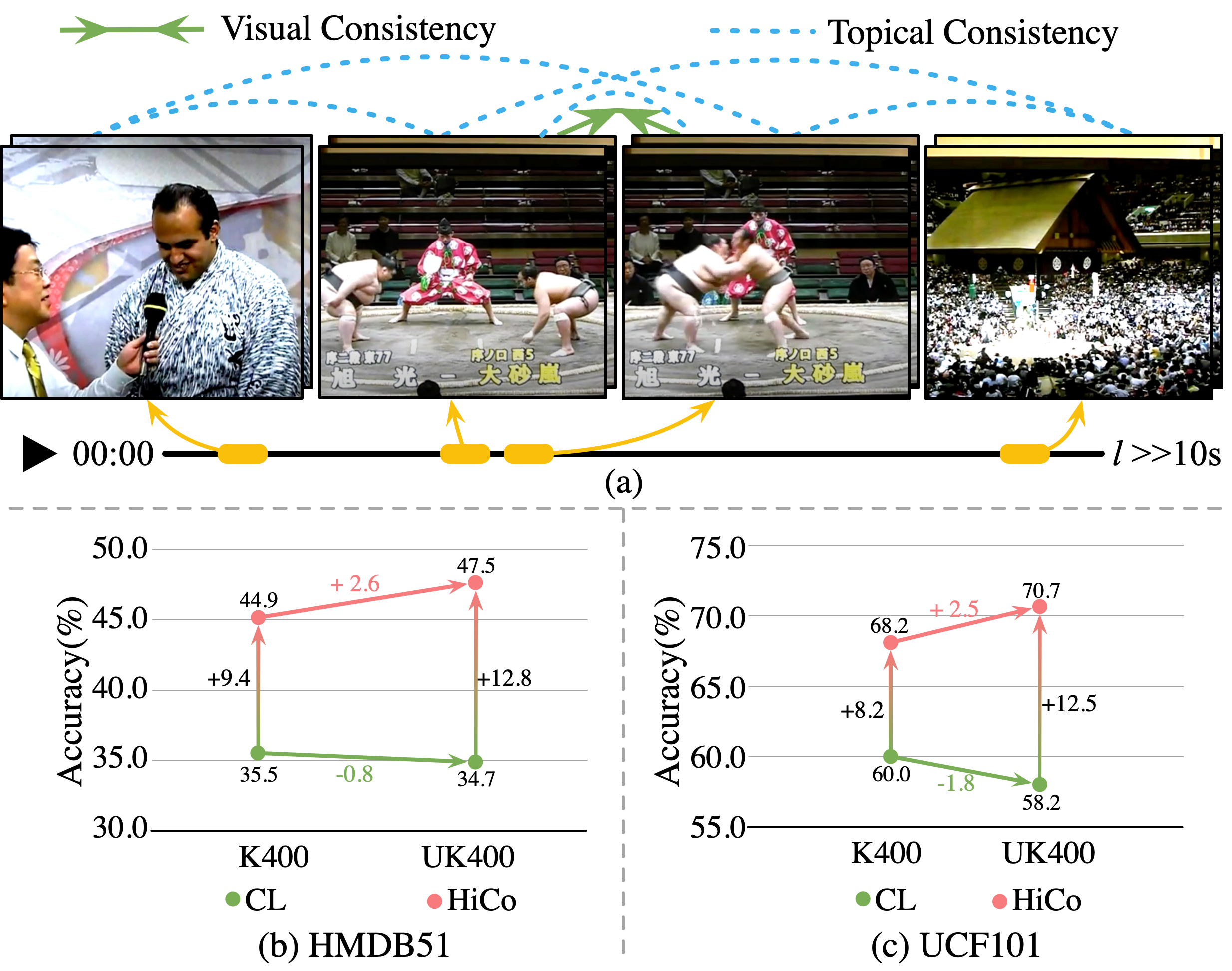}
    \vspace{-6mm}
    \caption{
    (a) An example of untrimmed video with the interview, competition and the stadium of \textit{Sumo Wrestling}. It shows the hierarchical consistency present in untrimmed videos.
    As can be seen, clips with short temporal distance share similar visual elements, while clips with long temporal distance, despite their dissimilar visual contents, share a same topic.
    (b, c) Linear evaluation of conventional contrastive learning (CL), \ie, SimCLR~\cite{chen2020simclr}, and HiCo on HMDB51~\cite{jhuang2011hmdb51} and UCF101~\cite{soomro2012ucf101}, with pretraining respectively on the original (trimmed) and untrimmed version of Kinetics-400~\cite{carreira2017k400}.
    }
    \label{fig:motivation}
    \vspace{-4mm}
\end{figure}

Self-supervised learning is of crucial importance in computer vision, and has shown remarkable potential in learning powerful spatio-temporal representations using unlabelled videos.
%
Current state-of-the-art approaches on unsupervised video representation learning are typically based on the contrastive learning framework~\cite{feichtenhofer2021largescale, qian2021cvrl, qing2021paramcrop},
which encourages the representations of the clips from the same video to be close and those from different videos to be as far away from each other as possible~\cite{chen2020simclr, he2020moco}.
In most approaches, they are trained on manually trimmed videos such as Kinetics-400~\cite{carreira2017k400}.
However, it is labor-intensive and time-consuming to collect such a large-scale trimmed video dataset, and the trimming process may also bring in certain human bias into the data.
In contrast, natural videos carry more abundant and diverse visual contents, and they are easier to obtain. 
Hence, this work sets out to exploit the natural \textit{untrimmed videos} for video representation learning.

%
Directly learning generalized and powerful representations from untrimmed videos is not a trivial problem, as empirical results both in Fig.~\ref{fig:motivation}(b, c) and in \cite{feichtenhofer2021largescale}(Tab.4 and Tab.6) demonstrate that directly applying contrastive learning on untrimmed videos yields \textit{worse} representations than on trimmed videos.
%
%
%
One possible reason is that the temporally-persistent hypothesis~\cite{feichtenhofer2021largescale} followed by the standard video contrastive learning framework and verified on trimmed videos is no longer sufficient for untrimmed videos.
Ideally, the temporally-persistent hypothesis learns an invariant representation for all the clips in the video.
This may be plausible for trimmed videos, and even for clips with short temporal distance in untrimmed ones, where a certain level of visual similarity or \textit{visual consistency} exists. 
Yet it could be \textit{overly strict} for temporally distant clips in untrimmed videos with less or no visual consistency exists, since they are only related by the same topic, \textit{i.e.,} they are \textit{topically consistent}.
In fact, we spot a hierarchical relation between the two consistencies existing in untrimmed videos.
Specifically, visually consistent pairs are always topically consistent, while topically consistent pairs are not necessarily visually consistent.
An example of the hierarchical consistency is visualized in Fig.~\ref{fig:motivation}(a).

In this paper, we present a novel framework for learning strong representations from \textit{untrimmed videos}.
By exploiting the hierarchical consistencies existing in untrimmed videos, \textit{i.e.,} the visual consistency and the topical consistency, our framework \textbf{HiCo} for Hierarchical Consistency learning can leverage the more abundant semantic patterns in natural videos.
We design two hierarchical tasks, respectively for learning the two consistencies. 
For visually consistent learning, we apply standard contrastive learning on clips with a small maximum temporal distance, and encourage temporally-invariant representations.
For topical consistency learning, we propose a topic prediction task, instead of a strict invariant mapping, the representations are only required to group different topics. 
Considering the hierarchical nature of consistencies, we also include the visually consistent pairs in topical consistency learning, while exclude topically consistent pairs for visual consistency learning.
%
%
%
Due to the complexity of the hierarchical tasks, we further introduce a gradual sampling that gradually increases the training difficulty for positive pairs to help optimization and improve generalization, which we show its superior both theoretically and empirically.
%
%

Extensive experiments on multiple downstream tasks show that employing HiCo can learn a strong and generalized video representation from untrimmed videos, with a convincing gap of 12.8\% and 12.5\% on the downstream action recognition task respectively on HMDB51~\cite{jhuang2011hmdb51} and UCF101~\cite{soomro2012ucf101} compared with the standard contrastive learning. We also demonstrate the capability of HiCo to learn a better representation from trimmed videos.

\section{Related Works}
\noindent
\textbf{Long video understanding.} 
The existing attempts for long video understanding is mainly based on supervised learning. 
\textit{Shot or event boundary detection} approaches~\cite{shao2015shot_det,baraldi2015shot_det,tang2018fast_shot_det,souvcek2019transnet_shot_det,gygli2018ridiculously_shot_det,shou2021long_form} aim to detect shot transitions or event boundaries in untrimmed videos. 
Among them, the former is caused by manual editing, and the latter is semantically-coherent. 
For \textit{temporal action localization}, existing works~\cite{qing2021tcanet,gao2020rapnet,lin2019bmn,lin2018bsn,xu2020gtad} attempt to distinguish action instances from unrelated complex backgrounds by modeling temporal relationships in untrimmed videos. 
Although the complex temporal structures in videos bring challenges for these tasks, there are many \textit{video classification} methods~\cite{wu2019long_lfb,tang2018non_long_vid_pre,yue2015beyond_long_vid_pre,li2017temporal_long_vid_pre,miech2017learnable_long_vid_pre,tang2020asynchronous_long_vid_pre,pang2021pgt,yang2021beyond} aggregate long-range temporal context to augment the predictions and achieve remarkable performance.
Unfortunately, these excellent supervised methods can not be transferred to self-supervised learning.
In this work, we try to leverage the inherent temporal structure in untrimmed videos for self-supervised video representation learning.

\begin{figure*}[h!]
\begin{center}
   \includegraphics[width=1.0\linewidth]{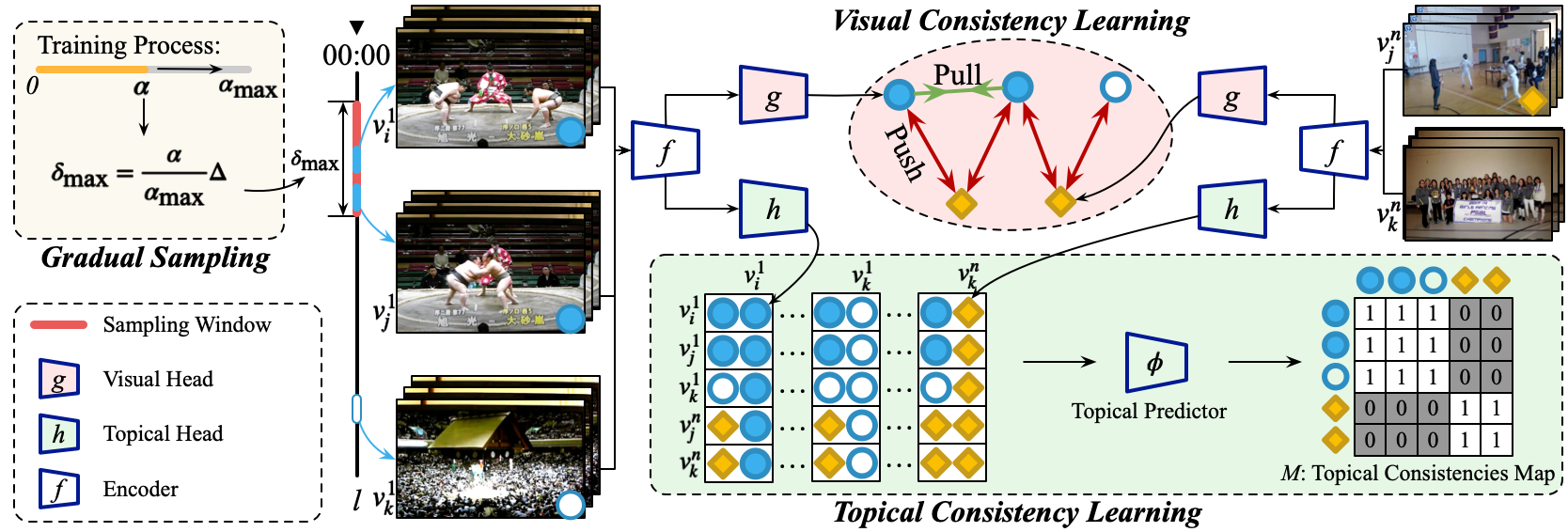}
\end{center}
\vspace{-4mm}
   \caption{The overall framework of HiCo. 
   HiCo contains three parts, including Visual Consistency Learning (VCL), Topical Consistency Learning (TCL), and Gradual Sampling (GS).
   VCL is based on standard contrastive learning to map a shared visual embedding for visually consistent pairs. TCL learns a topical predictor to discriminate the topical consistency between any two clips.
   The purpose of GS is to enhance both VCL and TCL by controlling the difficulty of training clips in each video.
   }
   \vspace{-3mm}
\label{fig:fw}
\end{figure*}

\noindent
\textbf{Self-supervised image representation learning.}
To avoid the labor-intensive annotation process, a wide range of self-supervised approaches have been proposed to exploit unlabeled data. 
Early methods mainly design different pretext tasks, including color restoration~\cite{zhang2016colorful}, image context reconstruction~\cite{pathak2016context_encoder} and solving jigsaw puzzles~\cite{noroozi2016jigsaw,doersch2015patch_pos}, \textit{etc}. 
Recently, contrastive learning based on instance discrimination has shown great potential in this field~\cite{chen2020simclr,he2020moco,chen2020mocov2,tian2019contrastive_endocing,wu2018ins_dis,oord2018cpc,henaff2020cpcv2}.
The main idea of contrastive learning is to train a transformation-invariant network. 

\noindent
\textbf{Self-supervised video representation learning.}
Existing self-supervised video representation learning approaches 
can be divide into three 
%
groups: designing different pretext tasks, applying contrastive learning and combining the both.
%
%
Pretext task based methods exploit the inherent structures 
%
naturally existing in videos to supervise the networks, such as speed perception~\cite{benaim2020speednet,yao2020prp,wang2020video_peace}, order prediction~\cite{misra2016shuffle_t_order,zhukov2020learning_t_order,fernando2017odd_one,lee2017unsupervised_t_order,xu2019vcop}, 
temporal transformation discrimination~\cite{jenni2020temporaltransformations}, motion estimation~\cite{wang2019self_motion, mosi}, and future prediction~\cite{diba2019dynamonet,srivastava2015lstm_unsupervised,luo2017unsupervised_future_predict,vondrick2016anticipating_future_predict,han2020memdpc}.
Contrastive learning related works~\cite{qian2021cvrl,qing2021paramcrop,feichtenhofer2021largescale,pan2021videomoco,ding2021fame} are mainly extended from image paradigm, 
%
and explore various spatio-temporal transformations for videos. 
It is worth noting that existing state-of-the-art methods are almost all based on the contrastive learning framework.
%
Further, there exist approaches to combine contrastive learning and temporal pretext tasks into a multi-task learning framework~\cite{kuang2021vcrl,jenni2021timeequ,bai2020can_multi_task,tao2020selfsupervised_muti_task}, which
enable the temporal exploration ability for contrastive learning and can further improve the video representations. 
%
%
%
%
Although previous methods have made significant progress for self-supervised video representation learning, they mostly rely on the curated videos that are manually trimmed beforehand and ignore the rich visual patterns embedded in original untrimmed videos.
In contrast, HiCo is a first attempt that focuses on self-supervised learning in untrimmed videos and enjoys both short-range and long-range temporal contexts simultaneously, as far as we know.
%
%

\section{Hierarchical Consistency Learning}
%
%
The main difference between the untrimmed videos and trimmed ones lies in the video length. 
For trimmed videos, any two random clips are likely to be visually similar since the temporal distance between two clips is always small.
However, for untrimmed videos, randomly sampled clip pairs could have a long temporal distance, making them only topically related, with low visual similarity.
On the other hand, clip pairs with short temporal distance could still be viewed as two clips sampled from trimmed videos, which share a high visual similarity. 
Hence, we divide the relations between the clip pairs into a hierarchy: 
\textbf{(i)} for clip pairs with short temporal distance with high visual similarity, we define their relationship as \textit{visually consistent;} 
\textbf{(ii)} for clip pairs with long temporal distance that may be only topic-related but visually dissimilar, we define their relationship as \textit{topically consistent.} 
Corresponding to this, we propose two hierarchical tasks to learn from the hierarchical consistencies, respectively visual consistency learning (VCL in Sec.~\ref{sec:vis_consistency}) and topical consistency learning (TCL in Sec.~\ref{sec:global_topic_method}).
%
%
%
%
%
%
%
%
%
%
Considering the complexity of the hierarchical tasks, we further propose a novel Gradual Sampling strategy to improve both VCL and TCL, and also provide theoretical analysis for its effectiveness.
Combined together, we present our overall framework HiCo in Fig.~\ref{fig:fw}.
%
%

\subsection{Visual Consistency Learning}
\label{sec:vis_consistency}
%

We learn visual consistency using the contrastive learning method SimCLR~\cite{chen2020simclr}. 
When applying contrastive learning for videos, it learns to map different clips from the same video (\ie, positive pairs) closer and repel clips from different videos (\ie, negative pairs).
Specifically, in a minibatch of $N$ videos, it samples two clips $v_i$ and $v_j$ from each video and thus generates $2N$ views with independent data augmentations.
After one latent vector $\mathbf{z}$ is extracted for each view through a backbone and a projection layer, the loss for the contrastive learning is formulated as:
%
%
%
%
%
%
\begin{equation}
    \mathcal{L}_{\text{CL}}=\frac{1}{2N}\sum_{n=1}^{N}[\ell(2n-1, 2n) + \ell(2n, 2n-1)] \ ,
\end{equation}
where $\ell(i, j)$ denotes the loss between two paired samples.
Given the cosine similarity $s_{i,j}$ between the representations $\mathbf{z}_i$ and $\mathbf{z}_j$, where $\{\mathbf{z}_i, \mathbf{z}_j\}=g(f({v_i, v_j}))$ with $f$ being the video backbone and $g$ being the contrastive projection head, $\ell(i, j)$ can be calculated as:
\begin{equation}
    \ell(i, j)=-\text{log}\frac{\text{exp}(s_{i,j} / \tau)}{\sum_{n=1}^{2N}\mathbbm{1}_{[n\ne i]}\text{exp}(s_{i,n}/\tau)}\ ,
\end{equation}
where $\tau$ represents the temperature and $\mathbbm{1}_{[n\ne i]}$ equals 1 if $n\ne i$, otherwise 0.

Since random sampling may yield $v_i$ and $v_j$ with low visual similarities in untrimmed videos, we further limit the maximum temporal distance for the clip pairs to learn the visual consistency. 
Formally, the temporal distance $\delta(v_i,v_j)$ between $v_i$ and $v_j$ is calculated and limited as:
\begin{equation}
    \delta(v_i,v_j) = |c_i-c_j| < \delta_{\text{max}}\ ,
\end{equation}
\noindent where $c_i$ and $c_j$ is the time step of the central frame in $v_i$ and $v_j$, and $\delta_{\text{max}}$ denotes the maximum distance between two sampled clips for visual consistency learning.
To guarantee the visual consistency between $v_i$ and $v_j$, $\delta_{\text{max}}$ should be significantly smaller than the video duration $l$, \textit{i.e.,} $0 \le \delta_{\text{max}} \ll l$.

\subsection{Topical Consistency Learning}
\label{sec:global_topic_method}
In general, the distant clips in untrimmed videos may be visually dissimilar but share the same topics, which is shown in the example in Fig.~\ref{fig:motivation} (a).
Although the scenes of interviews and stadium share little visual similarity with the competition, they all belong to the same topic of \emph{Sumo Wrestling}.
Hence, to fully exploit the visual diversities in untrimmed videos, we propose to learn from this topical consistency, which is overlooked in previous approaches. 

Formally, to learn topical consistency, we additionally randomly sample another clip $v_k$ from the entire video, which is not necessarily visually consistent to $v_i$ and $v_j$, but topically consistent to them.
However, due to the potential significant visual variations, it would be unreasonable for the topically consistent pairs to learn an invariant mapping. 
%
Therefore, we relax this strict constraint by (a) only introducing the $v_k$ as the negative sample for other videos in VCL; and (b) designing a learnable predictor to distinguish whether the input pairs are topically consistent, \textit{i.e.,} whether they belong to the same video.

With $v_k$ in the negative sample pool,
the loss between visually consistent pairs $\ell(i, j)$ are now calculated as follows: 
\begin{equation}
    \ell(i, j)=-\text{log}\frac{\text{exp}(s_{i,j} / \tau)}{\sum_{n=1}^{3N}\mathbbm{1}_{[n\ne i,k]}\text{exp}(s_{i,n}/\tau)}\ .
\end{equation}

%
%
For topic prediction, we first obtain the topical representations $\{\mathbf{t}_{i},\mathbf{t}_{j},\mathbf{t}_{k}\}$ for the sampled clips $\{v_{i},v_{j},v_{k}\}$ by the encoder $f(\cdot)$ and a topical project head $h(\cdot)$, \ie, $\{\mathbf{t}_{i},\mathbf{t}_{j},\mathbf{t}_{k}\} = h(f(\{v_{i},v_{j},v_{k}\}))$.
%
Given the $N$ videos with $3N$ clips in each mini-batch, the topical representations for all videos are combined to form a pair-wise feature set $\mathbf{U}$:
\begin{equation}
\small{\mathbf{U}=
    \left\{\begin{array}{ccccc}
    \mathbf{t}^1_{i}\oplus \mathbf{t}^1_{i}, & \mathbf{t}^1_{i}\oplus t^1_{j}, & \cdots & \mathbf{t}^1_{i}\oplus \mathbf{t}^N_{j}, & \mathbf{t}^1_{i}\oplus \mathbf{t}^N_{k},\\
    \vdots & \vdots & \ddots & \vdots & \vdots\\
    \mathbf{t}^N_{k}\oplus \mathbf{t}^1_{i}, & \mathbf{t}^N_{k}\oplus \mathbf{t}^1_{j}, & \cdots & \mathbf{t}^N_{k}\oplus \mathbf{t}^N_{j}, & \mathbf{t}^N_{k}\oplus \mathbf{t}^N_{k},\\
    \end{array}\right\},
}
\end{equation}
\noindent where the superscript $1...N$ denotes the video index, $\oplus$ denotes the concatenation and $\mathbf{U}\in\mathbb{R}^{3N\times 3N\times 2C_\text{T}}$ with $C_\text{T}$ being the dimension of the topical representation.
%
%
%
Finally, the topical consistencies $M$ for these pair-wise clips are estimated by a topical predictor:
%
\begin{equation}
    M=\phi(\mathbf{U}) \in R^{3N*3N}.
\end{equation}
\noindent
where the topical predictor $\phi(\cdot)$ is implemented by a Multi-Layer Perceptron (MLP). 
%
The supervised label for the topical consistencies $M$ is defined as $G\in R^{3N*3N}$, which indicates whether the pair-wise features share the same topic (\ie, whether they are from the same video).
During training, we apply focal loss~\cite{lin2017focalloss} $\mathcal{F}$ since the number of topically consistent pairs and inconsistent ones are heavily unbalanced.
The topic prediction loss is calculated as follows:

\begin{equation}
\small{
\mathcal{L}_\text{TP}=\frac{1}{\gamma_1}\sum_{G_{i,j}=1} \mathcal{F}(M_{i,j}) +\frac{1}{\gamma_2}\sum_{G_{i,j}=0} \mathcal{F}(1 - M_{i,j}),
}
\end{equation}
\noindent
where $\gamma_1$ and $\gamma_2$ are the number of positive samples and negative samples. 
%
Compared to visual consistency learning, where representations of the same video are encouraged to be identical, topical consistency learning poses a less strict constraint on the representations. 
%
Finally, the overall training objective of our HiCo is the sum of the contrastive loss and the topic prediction loss, formulated as $\mathcal{L}= \mathcal{L}_\text{CL}+ \mathcal{L}_\text{TP}$.
%
%


\subsection{Gradual Sampling}
\label{sec:cur_const}
%

%
Curriculum learning~\cite{bengio2009curriculum_learning} shows that models can learn much better when the training examples are not randomly provided but organized in a meaningful order, from easy examples to the hard ones.
It has achieved great success in a wide range of tasks.
Recalling that the untrimmed videos usually contain complex temporal contexts, randomly sampling clips unavoidablely generates dissimilar pairs in the early training stage, which can be considered as \emph{hard examples}
\footnote{For Visual Consistency Learning, even we limit the maximum temporal distance between clips $v_i$ and $v_j$, the training pairs are still considered \emph{harder} with a large temporal distance.}.
Therefore, we bring the spirit of curriculum learning into our HiCo, and propose a simple yet effective strategy to control the difficulty of positive pairs during the training stage, termed as Gradual Sampling.
%


Specifically, the $\delta_{\text{max}}$ is no longer a constant, but a function is driven by the current training epoch $\alpha$:
\begin{equation}\label{eqn:delta:alpha}
    \delta_{\text{max}}(\alpha) = \frac{\alpha}{\alpha_{\text{max}}}\Delta,
\end{equation}
where $\alpha$ and $\alpha_{\text{max}}$ refer to the current training epoch and the total training epoch, respectively. 
$\Delta$ is the upper bound of $\delta_{\text{max}}(\alpha)$, since $\alpha/\alpha_{\text{max}}$ satisfies the condition: $\alpha/\alpha_{\text{max}}\in[0,1]$, and here both $\alpha_{\text{max}}$ and $\Delta$ are constants.
%
%
This gradual sampling can be utilized to sample both visually consistent clips and topically consistent clips.

The $\delta_{\text{max}}(\alpha)$ linearly grows from $0$ to $\Delta$, which means we train the network from identical clips (with different data augmentations) and gradually increase the difficulty of positive pairs.
This can help improve the video representation generalization, and we will theoretically and experimentally show its superiority.
In fact, the gradual sampling in self-supervised learning can be also applied to both trimmed and  untrimmed videos.

%
%
%

\noindent
\textbf{Theoretical analysis.} 
We provide a theoretical understanding of the proposed gradual sampling strategy by leveraging the generalization analysis, which is common in the literature of learning theory~\cite{vapnik1999nature}. 
For the sake of analysis simplicity, we abstract the key points from the strategy, which are more math-friendly. 
To this end, we divide the training data into two groups, one with small variance (denoted by $\widehat\D_s$) and another one with large variance (denoted by $\widehat\D_l$). 
%
%
We let their population distributions as $\D_s$ and $\D_l$, respectively. 
Please note that this partition is for the proof use only, and it is not required in practice. 
At the early training epochs, the sampled clips are considered as examples with small variance since the sampling window size is small\footnote{When the window size is small, the sampled clips are usually similar.} according to Eq.\ref{eqn:delta:alpha}. 
While during the later training epochs, the sampled clips could be examples either with large or with small variance due to the larger sampling window size.
Let $\Lmath(w)$ be the loss function of the deep learning task that aims to be optimized, where $w$ is the model parameter. 
Given the output $\widehat w$ of an algorithm, the excess risk (ER) is a standard measure of generalization in learning theory~\cite{vapnik1999nature}, whose formulation is $\Lmath(\widehat w) - \Lmath(w_*)$, where $w_* = \arg\min_w \Lmath(w)$. 
The main goal is to obtain a solution $\widehat w$ as close as to the global optimal $w_*$. 
The following informal theorem (please refer to Appendix for its formal version) presents two excess risk bounds (ERB) to theoretically show why Gradual Sampling (GS) based sampling has better generalization than Random Sampling (RS) under some mild assumptions. 
Due to the space limitation, we include all other details, formal theorem, and proof in \textit{Appendix}.
\begin{thm}[Informal]\label{thm:main:informal}
Under some mild assumptions, the GS strategy can yield better generalization than the RS strategy.
Specifically, we have the following ERB in expectation: (1) for output of RS $\widehat  w_{rs}$, \begin{align*}
    \Lmath(\widehat w_{rs}) - \Lmath(w_*) \leq O\left( \Lmath(w_0) - \Lmath(w_*)\right);
\end{align*}
and (2) for output of GS $\widehat  w_{gs}$, 
\begin{align*}
    \Lmath(\widehat w_{gs}) - \Lmath(w_*) \leq O\left({\log(n)}/{n}+ p^2\hat{\Delta}^2\right),
\end{align*}
where $w_0$ is the initial solution, here $\hat{\Delta}$ is a measurement of difference between $\D_s$ and $\D_l$, $n$ is the sample size of $\widehat\D_s$ and $p\in[0,1]$ is the proportion of $\widehat\D_l$ among all training examples.
\end{thm}
The result (1) shows that RS did not receive a significant reduction in the objective due to the large variance arising from $D_l$.
On the other hand, the result (2) tells that GS could reduce the objective significantly when $n$ is large and $p$ is small, showing it has better generalization than RS. 
Please note that by appropriately selecting $\widehat\D_s$ and $\widehat\D_l$, $n$ can be large and $p$ is small enough, while theoretically the constant $\Lmath(w_0) - \Lmath(w_*)$ could be very large in general.

\section{Experiments}
\label{sec:exp}
\noindent
\textbf{Pre-training dataset.}
\textit{Kinetics-400}~\cite{carreira2017k400} \textit{(K400)} contains 240k trimmed videos, and each video lasts about 10 seconds. 
Since these short videos are trimmed from long videos, we recollect their original versions as our untrimmed video dataset, which we call \textit{untrimmed Kinetics-400 (UK400)}. 
Because many original videos are unavailable now, our UK400 dataset only contains 157k untrimmed videos for pre-training.
%
%
%
\textit{HACS}~\cite{zhao2019hacs} is a large-scale dataset for temporal action localization, which contains 37.6k long untrimmed videos for training. 
%
%

\noindent
\textbf{Pre-training settings.}
We choose SimCLR~\cite{chen2020simclr,qian2021cvrl} as basic contrastive learning framework, and adopt three frequently-used networks as encoder $f(\cdot)$, including S3D-G~\cite{xie2018s3dg}, R(2+1)D-10~\cite{tran2018r21d} and R3D-18~\cite{hara2018r3d}.
%
More training details about pre-training please refer to \textit{Appendix}.

\noindent
\textbf{Evaluations.}
%
We evaluate the representations learned by HiCo on three different downstream tasks, including action recognition, video retrieval and temporal action localization. 
Among them, action recognition and video retrieval are performed on two datasets: UCF101~\cite{soomro2012ucf101} and HMDB51~\cite{jhuang2011hmdb51}. 
For temporal action localization, we employ ActivityNet~\cite{caba2015anet} as evaluation dataset.
Please refer to \textit{Appendix} for more fine-tuning settings.
%
%
%

\noindent
\textbf{Note.}
In this section, unless otherwise specified, `FT/LFT' refers to fully fine-tuning/linear fine-tuning. 
`VCL', `TCL' and `GS' are Visual Consistency Learning, Topical Consistency Learning and Gradual Sampling, respectively.
Symbols `\cmark' and `\xmark' respectively indicate `Yes' and `No'.

\begin{table*}[t]
\centering
\begin{minipage}{0.75\linewidth}
\subfloat[
]{
\centering
\tablestyle{2pt}{1.05}
\begin{tabular}{c|ccc|cc}
\tabincell{c}{PT.}&VCL & TCL & GS & \tabincell{c}{HMDB51\\(\ft{FT}/LFT)} &  \tabincell{c}{UCF101\\(\ft{FT}/LFT)} \\
\shline
\multirow{4}*{HACS} &\xmark&\xmark&\xmark& \ft{42.9}/29.5  &  \ft{75.6}/54.9 \\  %
~                   &\cmark&\xmark&\xmark& \ft{42.7}/33.8  &  \ft{76.6}/57.9 \\  %
~                   &\xmark&\cmark&\xmark& \ft{42.6}/31.5  &  \ft{74.9}/55.9 \\ %
~                   &\cmark&\cmark&\xmark& \ft{43.9}/\textbf{35.6}  &  \ft{75.2}/\textbf{64.8} \\   
\hline
\multirow{4}*{UK400} &\xmark&\xmark&\xmark& \ft{45.1}/34.7 &  \ft{74.7}/58.2 \\   %
~                   &\cmark&\xmark&\xmark& \ft{47.9}/37.8  &  \ft{77.4}/65.2 \\  
~                    &\xmark&\cmark&\xmark& \ft{46.1}/34.8  &  \ft{77.2}/60.9 \\ 
~                    &\cmark&\cmark&\xmark& \ft{50.5}/\textbf{41.9}  &  \ft{77.7}/\textbf{68.8} \\  
\end{tabular}
}
\hspace{0.7em}
\subfloat[
]{
\centering
\footnotesize
\tablestyle{2pt}{1.05}
\begin{tabular}{c|ccc|cc}
\tabincell{c}{PT.}&VCL & TCL & GS & \tabincell{c}{HMDB51\\(\ft{FT}/LFT)} &  \tabincell{c}{UCF101\\(\ft{FT}/LFT)} \\
\shline
\multirow{4}*{HACS} &\xmark&\xmark&\cmark& \ft{43.8}/31.0  &  \ft{75.3}/57.4 \\ 
~                   &\cmark&\xmark&\cmark& \ft{48.7}/37.7  &  \ft{76.2}/63.0 \\   
~                   &\xmark&\cmark&\cmark& \ft{45.2}/32.0  &  \ft{75.3}/58.7 \\ 
~                   &\cmark&\cmark&\cmark& \ft{\textbf{51.8}}/\textbf{41.6}  &  \textbf{\ft{77.6}}/\textbf{67.6} \\ 
\hline
\multirow{4}*{UK400} &\xmark&\xmark&\cmark& \ft{46.1}/33.9  &  \ft{76.8}/59.3 \\ 
~                    &\cmark&\xmark&\cmark& \ft{51.2}/41.8  &  \ft{78.5}/67.2 \\  
~                    &\xmark&\cmark&\cmark& \ft{49.9}/36.1  &  \ft{76.7}/62.4 \\ 
~                    &\cmark&\cmark&\cmark& \ft{\textbf{54.1}}/\textbf{47.5}  &  \textbf{\ft{79.6}}/\textbf{70.7} \\ 
\end{tabular}
}
\end{minipage}
\hspace{-1.5em}
\begin{minipage}{0.25\linewidth}{\begin{center}
\subfloat[
]{
\tablestyle{2pt}{1.05}
    \begin{tabular}{c|cc}
    $\mathbf{U}$  & \tabincell{c}{HMDB51\\(\ft{FT}/LFT)} &  \tabincell{c}{UCF101\\(\ft{FT}/LFT)} \\
    \shline
    Uni. & \ft{52.4}/45.7  &  \ft{79.3}/69.0 \\
    Bi. & \ft{\textbf{54.1}}/\textbf{47.5}  &  \ft{\textbf{79.6}}/\textbf{70.7} \\
    \end{tabular}
}

\subfloat[
]{
\tablestyle{2pt}{1.05}
    \begin{tabular}{C{0.6cm}|cc}
    $v_k$  & \tabincell{c}{HMDB51\\(\ft{FT}/LFT)} &  \tabincell{c}{UCF101\\(\ft{FT}/LFT)} \\
    \shline
    \xmark  &\ft{48.8}/40.5  &  \ft{77.0}/68.0 \\  
    \cmark   &\ft{50.5}/\textbf{41.9}  &  \ft{77.7}/\textbf{68.8} \\  
    \end{tabular}
}
\end{center}
}\end{minipage}

\vspace{-3mm}
\caption{Ablation studies for HiCo with S3D-G. (a, b) Evaluating VCL and TCL both with and without GS. `PT.' refers to `Pre-training dataset'. (c) Bidirectional (Bi.) and unidirectional (Uni.) concatenation in $\mathbf{U}$. (d) Whether $v_k$ is in the negative sample pool of VCL.}
\vspace{-3mm}
\label{table:ablation_overall}
\end{table*}

\subsection{Ablation Study}

\begin{table}[t]
    \centering
\begin{minipage}{1.0\linewidth}{\begin{center}
\subfloat[
]{
\centering
\tablestyle{3pt}{1.05}
    \begin{tabular}{c|cc}
      \footnotesize $\Delta$(s)  & \tabincell{c}{HMDB51\\(\ft{FT}/LFT)} &  \tabincell{c}{UCF101\\(\ft{FT}/LFT)} \\
       \shline
        0.0          &\ft{51.9}/41.4 &\ft{80.4}/65.6 \\
        0.5          &\ft{54.9}/46.1 &\ft{79.5}/69.4 \\
        1.0          &\ft{54.1}/\textbf{47.5} &\ft{79.6}/\textbf{70.7} \\
        2.0          &\ft{51.7}/44.9 &\ft{77.5}/69.5 \\
        4.0          &\ft{52.2}/43.5 &\ft{79.6}/68.8 \\
    \end{tabular}
    \label{table:upper_bound_delta}
}
\hspace{0em}
\subfloat[
]{
\centering
\tablestyle{3pt}{1.05}
    \begin{tabular}{c|cc}
       \footnotesize Dis.(s)  & \tabincell{c}{HMDB51\\(\ft{FT}/LFT)} &  \tabincell{c}{UCF101\\(\ft{FT}/LFT)} \\
       \shline
        0                  &\ft{51.5}/43.6 &\ft{78.7}/66.8 \\
        10                 &\ft{53.3}/45.6 &\ft{80.1}/69.4 \\
        50                 &\ft{55.1}/45.1 &\ft{78.5}/69.7 \\
        100                &\ft{53.1}/45.8 &\ft{80.2}/70.3 \\
        $+\infty$          &\ft{54.1}/\textbf{47.5} &\ft{79.6}/\textbf{70.7} \\
    \end{tabular}
    \label{table:topical_pairs_dis}
}
\end{center}
}\end{minipage}
\vspace{-3mm}
\caption{Parameter sensitivity analysis. All experiments are conducted on UK400 with S3D-G. (a) The upper bound of $\delta_{\text{max}}(\alpha)$, \ie, $\Delta$. (b) The temporal distance of topical pairs.}
\label{table:pamater_analysis}
\end{table}

\noindent
\textbf{Importance of proposed VCL, TCL, and GS.}
%
%
Tab.~\ref{table:ablation_overall} ablates different components of HiCo.
%
%
%
%
From the results, we can find that:
first, VCL can improve standard contrastive learning on both datasets, which demonstrates visually consistent short-range clips can enhance the representation quality;
%
%
second, TCL alone is relatively weaker than VCL, but they are complementary to each other.
Combining VCL and TCL can respectively gain 7.2\% and 10.6\% on HMDB51 and UCF101 when pre-trained on UK400, as shown in Tab.~\ref{table:ablation_overall} (a);
%
%
%
third, GS can significantly improve VCL, TCL and their combination, especially it improves 5.6\% (41.9\% \emph{vs.} 47.5\%) on HMDB51 with UK400 pre-training.
%
%
%
Meanwhile, a similar trend is observed with HACS pre-training.
%
These results demonstrate the effectiveness of each component of HiCo.

\noindent
\textbf{Parameter sensitivity analysis for the upper bound of $\delta_{\text{max}}(\alpha)$, \ie, $\Delta$.}
Tab.~\ref{table:pamater_analysis}(a) presents results for varying $\Delta$.
It shows that the best performance is obtained when $\Delta$=1.0s. 
Note that, when $\Delta$ is set to 0s, the performances respectively drop 6.1\% and 5.1\% on HMDB51 and UCF101,
since all training examples are identical pairs without temporal variance, which declines the generalization.
Conversely, a large $\Delta$ may introduce dissimilar pairs with large visual variance, which may increase optimization difficulty and hence hurts the learned representations.

\begin{figure}
    \centering
    \includegraphics[width=1.0\linewidth]{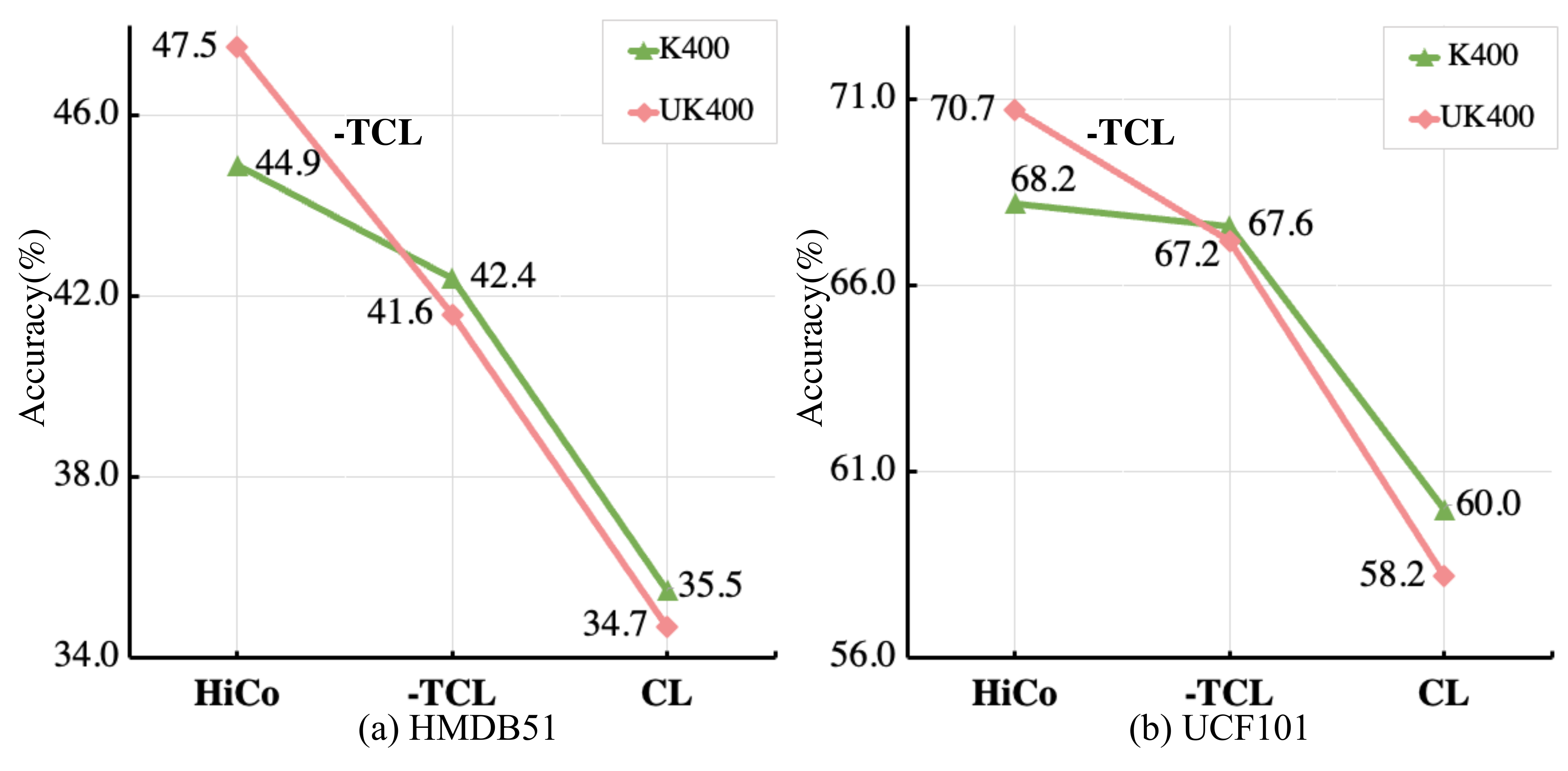}
    \vspace{-8mm}
    \caption{Removing TCL from HiCo. We pre-train S3D-G on K400 and UK400, and visualize the linear evaluations.}
    \vspace{-5mm}
    \label{fig:trimmed_abla}
\end{figure}

\begin{table}[t]
\centering
\tablestyle{5pt}{1.0}
\begin{tabular}{c|cc|cc}
\small Pretrain & \small Backbone  &  \small HiCo & \small \tabincell{c}{HMDB51\\(\ft{FT}/LFT)} & \small \tabincell{c}{UCF101\\(\ft{FT}/LFT)} \\
\shline
\multirow{6}*{HACS} & \multirow{2}*{S3D-G}  &\xmark & \ft{42.9}/29.5  &  \ft{75.6}/54.9 \\
~                   &  ~                   &\cmark &\ft{\textbf{51.8}}/\textbf{41.6}  &  \ft{\textbf{77.6}}/\textbf{67.6} \\
\cline{2-5}
~                   &\multirow{2}*{R(2+1)D-10}   &\xmark & \ft{47.7}/35.7 & \ft{81.3}/61.7 \\
~                   &~                     &\cmark &  \ft{\textbf{53.1}}/\textbf{43.7}  & \ft{\textbf{81.9}}/\textbf{71.3} \\
\cline{2-5}
~                   &\multirow{2}*{R3D-18}   &\xmark &  \ft{43.5}/32.8  &  \ft{72.8}/57.8 \\
~                   &~                      &\cmark &  \ft{\textbf{49.5}}/\textbf{43.3}  &  \ft{\textbf{76.1}}/\textbf{65.2} \\
\hline
\multirow{6}*{UK400} & \multirow{2}*{S3D-G}  &\xmark & \ft{45.1}/34.7 &  \ft{74.7}/58.2 \\
~                   &  ~                   &\cmark &\ft{\textbf{54.1}}/\textbf{47.5}  &  \ft{\textbf{79.6}}/\textbf{70.7} \\
\cline{2-5}
~                   &\multirow{2}*{R(2+1)D-10}   &\xmark &  \ft{47.4}/32.0  &  \ft{80.7}/57.4 \\
~                   &~                     &\cmark &  \ft{\textbf{50.9}}/\textbf{39.9}  &  \ft{\textbf{82.1}}/\textbf{67.7} \\
\cline{2-5}
~                   &\multirow{2}*{R3D-18}   &\xmark & \ft{44.4} /40.0  &  \ft{76.5}/65.5 \\
~                   &~                      &\cmark & \ft{ \textbf{47.7}}/\textbf{46.3}  &  \ft{\textbf{77.8}}/\textbf{70.7} \\
\hline
\multirow{2}*{K400} & \multirow{2}*{S3D-G}  &\xmark & \ft{46.2}/35.5 &  \ft{76.0}/60.0 \\
~                   &  ~                   &\cmark &\ft{\textbf{53.0}}/\textbf{44.9}  &  \ft{\textbf{79.0}}/\textbf{68.2} \\
\end{tabular}
\vspace{-3mm}
\caption{HiCo with different datasets and backbones.}
\vspace{-3mm}
\label{table:ablation_backbone}
\end{table}

\begin{table}[ht]
\centering
\tablestyle{5pt}{1.0}
\setlength{\tabcolsep}{1.3mm}{
\begin{tabular}{ccc|cc}
~ & \tabincell{c}{V.C.Pairs} &  \tabincell{c}{T.C.Pairs} & \tabincell{c}{HMDB51\\(\ft{FT}/LFT)} & \tabincell{c}{UCF101\\(\ft{FT}/LFT)} \\
\shline
(a) & $\mathcal{L}_{\text{CL}}$ & None  & \ft{47.9}/37.8  &  \ft{77.4}/65.2 \\
(b) & $\mathcal{L}_{\text{TP}}$ & None  & \ft{45.5}/19.7 & \ft{76.9}/27.3 \\   
(c) & $\mathcal{L}_{\text{CL}}$ & $\mathcal{L}_{\text{CL}}$   & \ft{46.9}/36.0 & \ft{76.1}/62.6 \\  
(d) & $\mathcal{L}_{\text{TP}}$ & $\mathcal{L}_{\text{TP}}$   & \ft{49.3}/24.7 & \ft{77.5}/38.9  \\ 
(e) & $\mathcal{L}_{\text{CL}}$+$\mathcal{L}_{\text{TP}}$ & $\mathcal{L}_{\text{TP}}$ & \ft{50.5}/41.9  &  \ft{77.7}/68.8 \\  
\hline
(f) & $\mathcal{L}_{\text{CL}}$+GS & None  & \ft{51.2}/41.8  &  \ft{78.5}/67.2 \\ 
(g) & $\mathcal{L}_{\text{TP}}$+GS & None  & \ft{47.6}/21.3 & \ft{77.0}/29.1 \\  
(h) & $\mathcal{L}_{\text{CL}}$+$\mathcal{L}_{\text{TP}}$+GS & None  & \ft{52.3}/43.5 & \ft{77.9}/65.7 \\ 
(i) & $\mathcal{L}_{\text{CL}}$+$\mathcal{L}_{\text{TP}}$ & $\mathcal{L}_{\text{TP}}$+GS & \ft{50.1}/41.9 & \ft{78.9}/69.7  \\ 
(j) & $\mathcal{L}_{\text{CL}}$+$\mathcal{L}_{\text{TP}}$+GS & $\mathcal{L}_{\text{TP}}$+GS  & \ft{\textbf{54.1}}/\textbf{47.5}  &  \ft{\textbf{79.6}}/\textbf{70.7} \\  

\end{tabular}}
\vspace{-3mm}
\caption{Ablation studies on loss functions. `V.C.Pairs' and `T.C.Pairs' are visually consistent pairs and topically consistent pairs, respectively.
}
\vspace{-4mm}
\label{table:ablation_global}
\end{table}

\begin{table*}[ht]
\centering
\tablestyle{5pt}{1.0}
\begin{tabular}{ccccccccc}
 Method &  Backbone &  Depth &  Pretrain   &   \tabincell{c}{PT Res.} &   \tabincell{c}{FT Res.} &  Freeze &  UCF101 &  HMDB51 \\
\shline
\textcolor{grey}{CVRL}~\cite{qian2021cvrl} & \textcolor{grey}{R3D} & \textcolor{grey}{50} & \textcolor{grey}{Kinetics-400} & \textcolor{grey}{$16\times224^2$} & \textcolor{grey}{$32\times224^2$} & \textcolor{grey}{\cmark} & \textcolor{grey}{89.8} & \textcolor{grey}{58.3} \\
CCL~\cite{kong2020ccl} & R3D & 18 & Kinetics-400 & $8\times112^2$ & $8\times112^2$ & \cmark & 52.1 & 27.8 \\
MLRep~\cite{qian2021mlrep} & R3D & 18 & Kinetics-400 & $16\times112^2$ & $16\times112^2$ & \cmark & 63.2 & 33.4 \\
FAME~\cite{ding2021fame} & R(2+1)D & 10 & Kinetics-400 & $16\times112^2$ & $16\times112^2$ & \cmark & 72.2 & 42.2 \\
CoCLR~\cite{han2020colr} & S3D & 23 & Kinetics-400 & $32\times128^2$ & $32\times128^2$ & \cmark & 74.5 & 46.1 \\
\hdashline
\textbf{HiCo(Ours)} & R3D & 18 & HACS & $8\times112^2$ & $16\times112^2$ & \cmark & 72.8 & 45.2 \\
\textbf{HiCo(Ours)} & S3D-G & 23 & Kinetics-400 & $16\times112^2$ & $16\times112^2$ & \cmark & 75.7 & 52.3 \\
\textbf{HiCo(Ours)} & R3D & 18 & UKinetics-400 & $8\times112^2$ & $16\times112^2$ & \cmark & 77.6 & 52.1 \\
\textbf{HiCo(Ours)} & R(2+1)D & 10 & Kinetics-400 & $16\times112^2$ & $16\times112^2$ & \cmark & 76.7 & 49.1 \\
\textbf{HiCo(Ours)} & R(2+1)D & 10 & UKinetics-400 & $16\times112^2$ & $16\times112^2$ & \cmark & \textbf{78.1} & 50.1 \\
\textbf{HiCo(Ours)} & S3D-G & 23 & UKinetics-400 & $16\times112^2$ & $16\times112^2$ & \cmark & 77.9 & \textbf{57.6} \\

\hline
\greyColor{VCLR}~\cite{kuang2021vcrl} & \greyColor{R2D} & \greyColor{50} & \greyColor{Kinetics-400} & \greyColor{$3\times224^2$} & \greyColor{$N/A\times224^2$} & \greyColor{\xmark} & \greyColor{85.6} & \greyColor{54.1} \\
\greyColor{$\rho$SimCLR}~\cite{feichtenhofer2021largescale} & \greyColor{R3D} & \greyColor{50} & \greyColor{Kinetics-400} & \greyColor{$8\times224^2$} & \greyColor{$8\times224^2$} & \greyColor{\xmark} & \greyColor{88.9} & \greyColor{-} \\
\greyColor{CVRL}~\cite{qian2021cvrl} & \greyColor{R3D} & \greyColor{50} & \greyColor{Kinetics-400} & \greyColor{$16\times224^2$} & \greyColor{$32\times224^2$} & \greyColor{\xmark} & \greyColor{92.2} & \greyColor{66.7} \\
\greyColor{$\rho$BYOL}~\cite{feichtenhofer2021largescale} & \greyColor{R3D} & \greyColor{50} & \greyColor{Kinetics-400} & \greyColor{$16\times224^2$} & \greyColor{$16\times224^2$} & \greyColor{\xmark} & \greyColor{95.5} & \greyColor{73.6} \\
VCLR~\cite{kuang2021vcrl} & R3D & 18 & HACS & $N/A\times224^2$ & $N/A\times224^2$ & \xmark & 67.2 & 49.3 \\
RSPNet~\cite{chen2020rspnet} & R3D & 18 & Kinetics-400 & $16\times112^2$ & $16\times112^2$ & \xmark & 81.1 & 44.6 \\
MLRep~\cite{qian2021mlrep} & R3D & 18 & Kinetics-400 & $16\times112^2$ & $16\times112^2$ & \xmark & 79.1 & 47.6 \\
ASCNet~\cite{huang2021ascnet} & R3D & 18 & Kinetics-400 & $16\times112^2$ & $16\times112^2$ & \xmark & 80.5 & 52.3 \\
VideoMoCo~\cite{pan2021videomoco} & R(2+1)D & 10 & Kinetics-400 & $32\times112^2$  & $32\times112^2$ & \xmark & 78.7 & 49.2 \\
SRTC~\cite{zhang2021srtc} & R(2+1)D & 10 & Kinetics-400 & $16\times112^2$ & $16\times112^2$ & \xmark & 82.0 & 51.2 \\
FAME~\cite{ding2021fame} & R(2+1)D & 10 & Kinetics-400 & $16\times112^2$ & $16\times112^2$ & \xmark & 84.8 & 53.5 \\
SpeedNet~\cite{benaim2020speednet} & S3D-G & 23 & Kinetics-400 & $64\times224^2$ & $64\times224^2$ & \xmark & 81.1 & 48.8 \\
RSPNet~\cite{chen2020rspnet} & S3D-G & 23 & Kinetics-400 & $64\times224^2$ & $64\times224^2$ & \xmark & 89.9 & 59.6 \\
\hdashline
\textbf{HiCo(Ours)} & R3D & 18 & HACS & $8\times112^2$ & $16\times112^2$ & \xmark & 77.0 & 56.2 \\
\textbf{HiCo(Ours)} & S3D-G & 23 & Kinetics-400 & $16\times112^2$ & $16\times112^2$ & \xmark & 83.2 & 56.3 \\
\textbf{HiCo(Ours)} & R3D & 18 & UKinetics-400 & $8\times112^2$ & $16\times112^2$ & \xmark & 87.2 & 63.7 \\
\textbf{HiCo(Ours)} & R(2+1)D & 10 & Kinetics-400 & $16\times112^2$ & $16\times112^2$ & \xmark & 85.3 & 57.9 \\
\textbf{HiCo(Ours)} & R(2+1)D & 10 & UKinetics-400 & $16\times112^2$ & $16\times112^2$ & \xmark & 86.5 & 55.6 \\
\textbf{HiCo(Ours)} & S3D-G & 23 & UKinetics-400 & $16\times112^2$ & $16\times112^2$ & \xmark & 83.6 & 60.4 \\
\textbf{HiCo(Ours)} & S3D-G & 23 & UKinetics-400 & $16\times112^2$ & $32\times224^2$ & \xmark & \textbf{91.0} & \textbf{66.5} \\

\end{tabular}
\vspace{-3mm}
\caption{Comparison to other state-of-the-art methods on the action recognition task. 
Where `Freeze' indicates freezing the parameters in backbones. 
`UKinetics-400' is untrimmed Kinetics-400 dataset. `PT Res.' and  `FT Res.' are sptial-temporal resolutions in pre-training and fine-tuning, respectively. 
`\greyColor{Grey fonts}' refers to the backbones different from HiCo.}
\vspace{-3mm}
\label{table:sota_action}
\end{table*}

\noindent
\textbf{Impact of the distance between topical pairs.}
Increasing the distance between topical pairs can introduce more temporal diversities, and the results under different settings are shown in Tab.~\ref{table:pamater_analysis}(b). 
We observe that the performance increases with a larger distance,
%
\eg, increasing the distance from 0 to $+\infty$ can lead to around 3.8\% gains on both datasets.
%
The results show that HiCo can effectively leverage the rich visual patterns from long-range topical pairs.

\noindent
\textbf{General applicabilities.}
%
Tab.~\ref{table:ablation_backbone} explores the generalities of HiCo with different backbones and datasets.
We can observe that HiCo can significantly boost performance, concerning both FT and LFT, from the baseline under all settings. 
%
%
Note that the performance of baseline pre-trained on UK400 is lower than that on K400, while HiCo can gain around 2.5\% with UK400 pre-training.
To further understand the reason behind this,
TCL is removed from HiCo in Fig.~\ref{fig:trimmed_abla}.
We can observe that the representations pre-trained on K400 are still stronger than UK400, similar to standard contrastive learning.
However, when integrating TCL, the performance pre-trained on UK400 surpasses K400 by 2.6\% on HMDB51, fully demonstrating that TCL can help to utilize the diverse temporal contexts in untrimmed videos to learn powerful representations.

\noindent
\textbf{Analysis of the loss function.}
Tab.~\ref{table:ablation_global} analyzes the hierarchical properties in HiCo from loss perspective.
We have several observations. 
\textit{(i)} Using $\mathcal{L}_{\text{TP}}$ alone is weaker than $\mathcal{L}_{\text{CL}}$. However, when combining $\mathcal{L}_{\text{TP}}$ and $\mathcal{L}_{\text{CL}}$, $\mathcal{L}_{\text{TP}}$ can further absorb the useful information from the topical pairs and improve the accuracy, \eg, 36.0\% \textit{vs.} 41.9\% on HMDB51, refer to (c, e). This shows the superiority of our proposed hierarchical learning architecture.
\textit{(ii)} From (b, d), we observe significant improvement by introducing topical pairs, which again confirms the importance of temporal diversities.
\textit{(iii)} From (a, f) and (b, g), separated $\mathcal{L}_{\text{CL}}$ and $\mathcal{L}_{\text{TP}}$ can be promoted by GS. Further comparing (e) and (j), even with extra topical pairs, GS can also strength the combined $\mathcal{L}_{\text{CL}}$ and $\mathcal{L}_{\text{TP}}$. Especially on HMDB51, integrating GS can improve accuracy by 5.6\%.
These experiments demonstrate the complementarity between visual and topical pairs and the effectiveness of GS from the loss perspective.

\begin{table*}[t]
\centering
\tablestyle{5pt}{1.0}
\begin{tabular}{ccccc|cccc|cccc}
\multirow{2}*{Method} &  \multirow{2}*{Backbone} &  \multirow{2}*{Depth} &  \multirow{2}*{Res.}  & \multirow{2}*{Pretrain}  & \multicolumn{4}{c}{UCF101} & \multicolumn{4}{c}{HMDB51} \\
 ~& ~ & ~ &  ~ ~ & & R@1 & R@5 & R@10 & R@20 & R@1 & R@5 & R@10 & R@20 \\
 \shline
\greyColor{VCLR}~\cite{kuang2021vcrl} & \greyColor{R2D}& \greyColor{50} & \greyColor{224} & \greyColor{K400}& \greyColor{70.6} & \greyColor{80.1} & \greyColor{86.3} & \greyColor{90.7} & \greyColor{35.2} & \greyColor{58.4}& \greyColor{68.8} & \greyColor{79.8}\\
RSPNet~\cite{chen2020rspnet} & R3D & 18 & 112 & K400 & 41.1 &  59.4 & 68.4 & 77.8 & - & - & - & - \\
MLRep~\cite{qian2021mlrep} & R3D & 18 & 112 & K400 & 41.5 &  60.0 & 71.2 & 80.1 & 20.7 & 40.8 & 55.2 & 68.3 \\
FAME~\cite{ding2021fame} &  R(2+1)D & 10 & 112 & K400 & 62.3 &  75.1 & 80.9 & 86.9 & - & - & - & - \\
SpeedNet~\cite{benaim2020speednet} & S3D-G & 23 & 224 & K400 & 13.0 &  28.1 & 37.5 & 49.5 & - & - & - & - \\
\hdashline
\textbf{HiCo(Ours)} & R3D & 18 & 112& UK400 & \textbf{71.8}&\textbf{83.8}&88.5&92.8 & \textbf{35.8}&59.7&71.1&81.2 \\
\textbf{HiCo(Ours)} & R(2+1)D & 10 & 112& UK400 & 69.1&84.4&\textbf{89.0}&\textbf{93.6} & 35.2&58.8&70.3&\textbf{82.3} \\
\textbf{HiCo(Ours)} & S3D-G & 23 & 112& UK400 & 62.5   & 76.4 & 82.9 & 89.4 & 35.5 & \textbf{60.3} & \textbf{72.2} & 82.1 \\
\end{tabular}
\vspace{-3mm}
\caption{Nearest neighobor retrieval comparison on UCF101 and HMDB51. `\greyColor{Grey fonts}' refer different backbones with HiCo.}
\vspace{-3mm}
\label{table:sota_retrieval}
\end{table*}

\begin{table}[t]
    \centering
    \small
    \begin{tabular}{ccc|cc}
      Method & Encoder & PT Data. & \tabincell{c}{AUC}   & \tabincell{c}{AR@100} \\
      \shline
      VINCE~\cite{gordon2020vince} & R2D-50 &   K400  &  64.6\%  & 73.2\% \\
      SeCo~\cite{yao2020seco} & R2D-50 &   K400  &  65.2\%  & 73.4\% \\
      VCLR~\cite{kuang2021vcrl} & R2D-50 &   K400  &  65.5\%  & 73.8\% \\
      \hline
      CL &S3D-G  &   UK400 &  63.0\%  &  72.4\% \\
      HiCo &S3D-G  &   UK400 &  \textbf{67.1\%}  & \textbf{75.4\%} \\
    \end{tabular}
    \vspace{-3mm}
    \caption{Action localization on ActivityNet~\cite{caba2015anet}. `PT Data.' refers to pre-training dataset.}
    \vspace{-3mm}
    \label{tab:localization}
\end{table}

\noindent
\textbf{More explorations.}
\textit{(i)} Tab.~\ref{table:ablation_overall}(c) reports two different concatenating ways for pair-wise topical features in the feature set $\mathbf{U}$. 
The unidirectional one shows weaker performance, since the bidirectional setting can provide more expert prior; that is, topical consistencies are unrelated to feature orders.
\textit{(ii)} Tab.~\ref{table:ablation_overall}(d) explores the necessity of incorporating $v_k$ into the negative pool for visually consistent pairs (\ie $v_i$ and $v_j$). Although $v_k$ may be visually dissimilar with $v_i$ and $v_j$, it can also provide extra supervision signals for VCL and improve generalization.

\subsection{Evaluation on action recognition task}
Tab.~\ref{table:sota_action} compares HiCo with other state-of-the-art methods.
We list the relevant settings in details for fair comparisons, including network architectures and training resolutions.
From the table, we draw the following conclusions.
First, in terms of linear evaluation, HiCo notably outperforms existing methods under similar settings.
HiCo surpasses CoCLR~\cite{han2020colr} by 3.4\% and 11.5\% on UCF101 and HMDB51, respectively, even they use more frames and optical flow in pre-training.
%
%
Due to the use of a deeper network and larger resolution in CVRL~\cite{qian2021cvrl}, HiCo achieves a slightly lower performance, but the gap is significantly closed when fully fine-tuning is employed.
%
%
%
Second, when pre-training without freezing backbones, 
HiCo achieves better performances than previous approaches under similar settings. 
For example, using same input resolutions ($16\times 112^2$) and backbone (R(2+1)D), HiCo is 1.7\% and 2.1\% higher than FAME~\cite{ding2021fame} on UCF101 and HMDB51, respectively.
%
%
Note that $\rho$BYOL~\cite{feichtenhofer2021largescale} obtains excellent performance. The reason may be that it applies a different self-supervised learning method (BYOL), deeper network, and large resolution.
When adopting the same SimCLR, HiCo can achieve comparable performance with $\rho$SimCLR~\cite{feichtenhofer2021largescale}, using lower resolutions and a tinyer backbone.
Third, compared to VCLR~\cite{kuang2021vcrl} trained on HACS using R3D-18, a notable improvement is observed with similar conditions on both datasets, with a gap of 9.8\% and 6.9\% on respective datasets. 
This demonstrates that HiCo is a more suitable framework for learning from untrimmed videos.

\subsection{Evaluation on video retrieval task}
%
We compute normalized cosine similarity with features extracted by UK400 pre-trained networks for video retrieval.
%
%
Tab.~\ref{table:sota_retrieval} compares HiCo with other approaches with different top-\textit{k} accuracies.
HiCo exceeds the state-of-the-art method (\ie, VCLR~\cite{kuang2021vcrl}) by 1.2\% on UCF101 under R@1 with a lightweight R3D-18 network, 
which implies that features learned by HiCo is more generalized.

\subsection{Evaluation on action localization task}
%
%
We use the mainstream TAL method, \ie, BMN~\cite{lin2019bmn}, to evaluate the UK400 pre-trained features on ActivityNet~\cite{caba2015anet}.
%
%
%
%
As shown in Tab.~\ref{tab:localization}, HiCo with a lightweight encoder significantly outperforms VCLR~\cite{kuang2021vcrl} by 1.6\% in terms of AUC. Compared with standard contrastive learning, HiCo boosts the AUC by 4.1\%.
%
%
The main reason is that HiCo can preserve more high-level information for clips through topically consistent learning, which assists BMN in discriminating the actions and background.
The results successfully demonstrate the transferability of HiCo pre-trained representations to different downstream tasks.

\section{Discussions}
\noindent
\textbf{Limitations.} 
HiCo provides a simple framework for learning representations in untrimmed videos. 
Despite its effectiveness on existing public datasets, HiCo may fail when it encounters videos with various topics and more complex relations between different clips, such as movies or TV series. 
Therefore, for learning from unlabelled untrimmed videos, one should avoid using videos sampled from those sources. 

%

\noindent
\textbf{Conclusion. }
In this paper, we propose HiCo, a novel self-supervised learning framework for learning powerful video representations from untrimmed videos. 
It exploits the hierarchical consistency existing in the long videos, \textit{i.e.,} the visual consistency and the topical consistency. 
For visual consistency learning, HiCo employs the contrastive learning with constrained clip distances. 
For topical consistency learning, a topic prediction task is presented.
Further, a gradual sampling strategy is proposed based on curriculum learning for both tasks, whose superiority is demonstrated theoretically and empirically.
In general, we believe that untrimmed videos are not only easier to collect, but also provide more potential for learning more robust video representations.
We hope the simple and effective framework of HiCo can encourage and inspire the researchers to be further devoted to this area.

\noindent\textbf{Acknowledgement. }This work is supported by the National Natural Science Foundation of China under grant 61871435, Fundamental Research Funds for the Central Universities no.2019kfyXKJC024, by the 111 Project on Computational Intelligence and Intelligent Control under Grant B18024, by the Agency for Science, Technology and Research (A*STAR) under its AME Programmatic Funding Scheme (Project \#A18A2b0046) and by Alibaba Group through Alibaba Research Intern Program.
%
%
%
%
%

{\small
\bibliographystyle{ieee_fullname}
\bibliography{egbib}

\begin{thebibliography}{10}\itemsep=-1pt

\bibitem{bai2020can_multi_task}
Yutong Bai, Haoqi Fan, Ishan Misra, Ganesh Venkatesh, Yongyi Lu, Yuyin Zhou,
  Qihang Yu, Vikas Chandra, and Alan Yuille.
\newblock Can temporal information help with contrastive self-supervised
  learning?
\newblock {\em arXiv preprint arXiv:2011.13046}, 2020.

\bibitem{baraldi2015shot_det}
Lorenzo Baraldi, Costantino Grana, and Rita Cucchiara.
\newblock Shot and scene detection via hierarchical clustering for re-using
  broadcast video.
\newblock In {\em International Conference on Computer Analysis of Images and
  Patterns}, pages 801--811. Springer, 2015.

\bibitem{benaim2020speednet}
Sagie Benaim, Ariel Ephrat, Oran Lang, Inbar Mosseri, William~T Freeman,
  Michael Rubinstein, Michal Irani, and Tali Dekel.
\newblock Speednet: Learning the speediness in videos.
\newblock In {\em CVPR}, pages 9922--9931, 2020.

\bibitem{bengio2009curriculum_learning}
Yoshua Bengio, J{\'e}r{\^o}me Louradour, Ronan Collobert, and Jason Weston.
\newblock Curriculum learning.
\newblock In {\em ICML}, pages 41--48, 2009.

\bibitem{bertsekas1995neuro}
Dimitri~P Bertsekas and John~N Tsitsiklis.
\newblock Neuro-dynamic programming: an overview.
\newblock In {\em Proceedings of 1995 34th IEEE Conference on Decision and
  Control}, volume~1, pages 560--564. IEEE, 1995.

\bibitem{bottou2018optimization}
L{\'e}on Bottou, Frank~E Curtis, and Jorge Nocedal.
\newblock Optimization methods for large-scale machine learning.
\newblock {\em Siam Review}, 60(2):223--311, 2018.

\bibitem{carreira2017k400}
Joao Carreira and Andrew Zisserman.
\newblock Quo vadis, action recognition? a new model and the kinetics dataset.
\newblock In {\em CVPR}, pages 6299--6308, 2017.

\bibitem{chen2020rspnet}
Peihao Chen, Deng Huang, Dongliang He, Xiang Long, Runhao Zeng, Shilei Wen,
  Mingkui Tan, and Chuang Gan.
\newblock Rspnet: Relative speed perception for unsupervised video
  representation learning.
\newblock {\em arXiv preprint arXiv:2011.07949}, 2020.

\bibitem{chen2020simclr}
Ting Chen, Simon Kornblith, Mohammad Norouzi, and Geoffrey Hinton.
\newblock A simple framework for contrastive learning of visual
  representations.
\newblock In {\em ICML}, pages 1597--1607. PMLR, 2020.

\bibitem{chen2020mocov2}
Xinlei Chen, Haoqi Fan, Ross Girshick, and Kaiming He.
\newblock Improved baselines with momentum contrastive learning.
\newblock {\em arXiv preprint arXiv:2003.04297}, 2020.

\bibitem{diba2019dynamonet}
Ali Diba, Vivek Sharma, Luc~Van Gool, and Rainer Stiefelhagen.
\newblock Dynamonet: Dynamic action and motion network.
\newblock In {\em ICCV}, pages 6192--6201, 2019.

\bibitem{ding2021fame}
Shuangrui Ding, Maomao Li, Tianyu Yang, Rui Qian, Haohang Xu, Qingyi Chen, and
  Jue Wang.
\newblock Motion-aware self-supervised video representation learning via
  foreground-background merging.
\newblock {\em arXiv preprint arXiv:2109.15130}, 2021.

\bibitem{doersch2015patch_pos}
Carl Doersch, Abhinav Gupta, and Alexei~A Efros.
\newblock Unsupervised visual representation learning by context prediction.
\newblock In {\em ICCV}, pages 1422--1430, 2015.

\bibitem{caba2015anet}
Bernard~Ghanem Fabian Caba~Heilbron, Victor~Escorcia and Juan~Carlos Niebles.
\newblock Activitynet: A large-scale video benchmark for human activity
  understanding.
\newblock In {\em Proceedings of the IEEE Conference on Computer Vision and
  Pattern Recognition}, pages 961--970, 2015.

\bibitem{feichtenhofer2021largescale}
Christoph Feichtenhofer, Haoqi Fan, Bo Xiong, Ross Girshick, and Kaiming He.
\newblock A large-scale study on unsupervised spatiotemporal representation
  learning.
\newblock In {\em Proceedings of the IEEE/CVF Conference on Computer Vision and
  Pattern Recognition}, pages 3299--3309, 2021.

\bibitem{fernando2017odd_one}
Basura Fernando, Hakan Bilen, Efstratios Gavves, and Stephen Gould.
\newblock Self-supervised video representation learning with odd-one-out
  networks.
\newblock In {\em CVPR}, pages 3636--3645, 2017.

\bibitem{gao2020rapnet}
Jialin Gao, Zhixiang Shi, Guanshuo Wang, Jiani Li, Yufeng Yuan, Shiming Ge, and
  Xi Zhou.
\newblock Accurate temporal action proposal generation with relation-aware
  pyramid network.
\newblock In {\em AAAI}, pages 10810--10817, 2020.

\bibitem{ghadimi2013stochastic}
Saeed Ghadimi and Guanghui Lan.
\newblock Stochastic first-and zeroth-order methods for nonconvex stochastic
  programming.
\newblock {\em SIAM Journal on Optimization}, 23(4):2341--2368, 2013.

\bibitem{gordon2020vince}
Daniel Gordon, Kiana Ehsani, Dieter Fox, and Ali Farhadi.
\newblock Watching the world go by: Representation learning from unlabeled
  videos.
\newblock {\em arXiv preprint arXiv:2003.07990}, 2020.

\bibitem{gygli2018ridiculously_shot_det}
Michael Gygli.
\newblock Ridiculously fast shot boundary detection with fully convolutional
  neural networks.
\newblock In {\em 2018 International Conference on Content-Based Multimedia
  Indexing (CBMI)}, pages 1--4. IEEE, 2018.

\bibitem{han2020memdpc}
Tengda Han, Weidi Xie, and Andrew Zisserman.
\newblock Memory-augmented dense predictive coding for video representation
  learning.
\newblock {\em arXiv preprint arXiv:2008.01065}, 2020.

\bibitem{han2020colr}
Tengda Han, Weidi Xie, and Andrew Zisserman.
\newblock Self-supervised co-training for video representation learning.
\newblock {\em arXiv preprint arXiv:2010.09709}, 2020.

\bibitem{hara2018r3d}
Kensho Hara, Hirokatsu Kataoka, and Yutaka Satoh.
\newblock Can spatiotemporal 3d cnns retrace the history of 2d cnns and
  imagenet?
\newblock In {\em CVPR}, pages 6546--6555, 2018.

\bibitem{he2020moco}
Kaiming He, Haoqi Fan, Yuxin Wu, Saining Xie, and Ross Girshick.
\newblock Momentum contrast for unsupervised visual representation learning.
\newblock In {\em CVPR}, pages 9729--9738, 2020.

\bibitem{henaff2020cpcv2}
Olivier Henaff.
\newblock Data-efficient image recognition with contrastive predictive coding.
\newblock In {\em International Conference on Machine Learning}, pages
  4182--4192. PMLR, 2020.

\bibitem{huang2021ascnet}
Deng Huang, Wenhao Wu, Weiwen Hu, Xu Liu, Dongliang He, Zhihua Wu, Xiangmiao
  Wu, Mingkui Tan, and Errui Ding.
\newblock Ascnet: Self-supervised video representation learning with
  appearance-speed consistency.
\newblock {\em arXiv preprint arXiv:2106.02342}, 2021.

\bibitem{mosi}
Ziyuan Huang, Shiwei Zhang, Jianwen Jiang, Mingqian Tang, Rong Jin, and
  Marcelo~H Ang.
\newblock Self-supervised motion learning from static images.
\newblock In {\em Proceedings of the IEEE/CVF Conference on Computer Vision and
  Pattern Recognition}, pages 1276--1285, 2021.

\bibitem{jenni2021timeequ}
Simon Jenni and Hailin Jin.
\newblock Time-equivariant contrastive video representation learning.
\newblock In {\em Proceedings of the IEEE/CVF International Conference on
  Computer Vision}, pages 9970--9980, 2021.

\bibitem{jenni2020temporaltransformations}
Simon Jenni, Givi Meishvili, and Paolo Favaro.
\newblock Video representation learning by recognizing temporal
  transformations.
\newblock In {\em Computer Vision--ECCV 2020: 16th European Conference,
  Glasgow, UK, August 23--28, 2020, Proceedings, Part XXVIII 16}, pages
  425--442. Springer, 2020.

\bibitem{jhuang2011hmdb51}
H Jhuang, H Garrote, E Poggio, T Serre, and T Hmdb.
\newblock A large video database for human motion recognition.
\newblock In {\em ICCV}, volume~4, page~6, 2011.

\bibitem{kingma2014adam}
Diederik~P Kingma and Jimmy Ba.
\newblock Adam: A method for stochastic optimization.
\newblock {\em arXiv preprint arXiv:1412.6980}, 2014.

\bibitem{kong2020ccl}
Quan Kong, Wenpeng Wei, Ziwei Deng, Tomoaki Yoshinaga, and Tomokazu Murakami.
\newblock Cycle-contrast for self-supervised video representation learning.
\newblock {\em arXiv preprint arXiv:2010.14810}, 2020.

\bibitem{kuang2021vcrl}
Haofei Kuang, Yi Zhu, Zhi Zhang, Xinyu Li, Joseph Tighe, Soren Schwertfeger,
  Cyrill Stachniss, and Mu Li.
\newblock Video contrastive learning with global context.
\newblock In {\em Proceedings of the IEEE/CVF International Conference on
  Computer Vision}, pages 3195--3204, 2021.

\bibitem{lee2017unsupervised_t_order}
Hsin-Ying Lee, Jia-Bin Huang, Maneesh Singh, and Ming-Hsuan Yang.
\newblock Unsupervised representation learning by sorting sequences.
\newblock In {\em Proceedings of the IEEE International Conference on Computer
  Vision}, pages 667--676, 2017.

\bibitem{li2017temporal_long_vid_pre}
Fu Li, Chuang Gan, Xiao Liu, Yunlong Bian, Xiang Long, Yandong Li, Zhichao Li,
  Jie Zhou, and Shilei Wen.
\newblock Temporal modeling approaches for large-scale youtube-8m video
  understanding.
\newblock {\em arXiv preprint arXiv:1707.04555}, 2017.

\bibitem{lin2019bmn}
Tianwei Lin, Xiao Liu, Xin Li, Errui Ding, and Shilei Wen.
\newblock Bmn: Boundary-matching network for temporal action proposal
  generation.
\newblock In {\em Proceedings of the IEEE International Conference on Computer
  Vision}, pages 3889--3898, 2019.

\bibitem{lin2018bsn}
Tianwei Lin, Xu Zhao, Haisheng Su, Chongjing Wang, and Ming Yang.
\newblock Bsn: Boundary sensitive network for temporal action proposal
  generation.
\newblock In {\em Proceedings of the European Conference on Computer Vision
  (ECCV)}, pages 3--19, 2018.

\bibitem{lin2017focalloss}
Tsung-Yi Lin, Priya Goyal, Ross Girshick, Kaiming He, and Piotr Doll{\'a}r.
\newblock Focal loss for dense object detection.
\newblock In {\em Proceedings of the IEEE international conference on computer
  vision}, pages 2980--2988, 2017.

\bibitem{loshchilov2017adamw}
Ilya Loshchilov and Frank Hutter.
\newblock Decoupled weight decay regularization.
\newblock {\em arXiv preprint arXiv:1711.05101}, 2017.

\bibitem{luo2017unsupervised_future_predict}
Zelun Luo, Boya Peng, De-An Huang, Alexandre Alahi, and Li Fei-Fei.
\newblock Unsupervised learning of long-term motion dynamics for videos.
\newblock In {\em Proceedings of the IEEE conference on computer vision and
  pattern recognition}, pages 2203--2212, 2017.

\bibitem{miech2017learnable_long_vid_pre}
Antoine Miech, Ivan Laptev, and Josef Sivic.
\newblock Learnable pooling with context gating for video classification.
\newblock {\em arXiv preprint arXiv:1706.06905}, 2017.

\bibitem{misra2016shuffle_t_order}
Ishan Misra, C~Lawrence Zitnick, and Martial Hebert.
\newblock Shuffle and learn: unsupervised learning using temporal order
  verification.
\newblock In {\em European Conference on Computer Vision}, pages 527--544.
  Springer, 2016.

\bibitem{opac-b1104789}
Yurii Nesterov.
\newblock {\em Introductory lectures on convex optimization : a basic course}.
\newblock Applied optimization. Kluwer Academic Publ., 2004.

\bibitem{noroozi2016jigsaw}
Mehdi Noroozi and Paolo Favaro.
\newblock Unsupervised learning of visual representations by solving jigsaw
  puzzles.
\newblock In {\em ECCV}, pages 69--84. Springer, 2016.

\bibitem{oord2018cpc}
Aaron van~den Oord, Yazhe Li, and Oriol Vinyals.
\newblock Representation learning with contrastive predictive coding.
\newblock {\em arXiv preprint arXiv:1807.03748}, 2018.

\bibitem{pan2021videomoco}
Tian Pan, Yibing Song, Tianyu Yang, Wenhao Jiang, and Wei Liu.
\newblock Videomoco: Contrastive video representation learning with temporally
  adversarial examples.
\newblock In {\em Proceedings of the IEEE/CVF Conference on Computer Vision and
  Pattern Recognition}, pages 11205--11214, 2021.

\bibitem{pang2021pgt}
Bo Pang, Gao Peng, Yizhuo Li, and Cewu Lu.
\newblock Pgt: A progressive method for training models on long videos.
\newblock In {\em Proceedings of the IEEE/CVF Conference on Computer Vision and
  Pattern Recognition}, pages 11379--11389, 2021.

\bibitem{pathak2016context_encoder}
Deepak Pathak, Philipp Krahenbuhl, Jeff Donahue, Trevor Darrell, and Alexei~A
  Efros.
\newblock Context encoders: Feature learning by inpainting.
\newblock In {\em CVPR}, pages 2536--2544, 2016.

\bibitem{polyak1963gradient}
Boris~Teodorovich Polyak.
\newblock Gradient methods for minimizing functionals.
\newblock {\em Zhurnal Vychislitel'noi Matematiki i Matematicheskoi Fiziki},
  3(4):643--653, 1963.

\bibitem{qian2021mlrep}
Rui Qian, Yuxi Li, Huabin Liu, John See, Shuangrui Ding, Xian Liu, Dian Li, and
  Weiyao Lin.
\newblock Enhancing self-supervised video representation learning via
  multi-level feature optimization.
\newblock In {\em Proceedings of the IEEE/CVF International Conference on
  Computer Vision}, pages 7990--8001, 2021.

\bibitem{qian2021cvrl}
Rui Qian, Tianjian Meng, Boqing Gong, Ming-Hsuan Yang, Huisheng Wang, Serge
  Belongie, and Yin Cui.
\newblock Spatiotemporal contrastive video representation learning.
\newblock In {\em Proceedings of the IEEE/CVF Conference on Computer Vision and
  Pattern Recognition}, pages 6964--6974, 2021.

\bibitem{qing2021paramcrop}
Zhiwu Qing, Ziyuan Huang, Shiwei Zhang, Mingqian Tang, Changxin Gao, Marcelo~H
  Ang~Jr, Rong Jin, and Nong Sang.
\newblock Paramcrop: Parametric cubic cropping for video contrastive learning.
\newblock {\em arXiv preprint arXiv:2108.10501}, 2021.

\bibitem{qing2021tcanet}
Zhiwu Qing, Haisheng Su, Weihao Gan, Dongliang Wang, Wei Wu, Xiang Wang, Yu
  Qiao, Junjie Yan, Changxin Gao, and Nong Sang.
\newblock Temporal context aggregation network for temporal action proposal
  refinement.
\newblock In {\em Proceedings of the IEEE/CVF Conference on Computer Vision and
  Pattern Recognition}, pages 485--494, 2021.

\bibitem{shao2015shot_det}
Hong Shao, Yang Qu, and Wencheng Cui.
\newblock Shot boundary detection algorithm based on hsv histogram and hog
  feature.
\newblock In {\em 5th International Conference on Advanced Engineering
  Materials and Technology}, pages 951--957, 2015.

\bibitem{shou2021long_form}
Mike~Zheng Shou, Stan~W Lei, Weiyao Wang, Deepti Ghadiyaram, and Matt Feiszli.
\newblock Generic event boundary detection: A benchmark for event segmentation.
\newblock {\em arXiv preprint arXiv:2101.10511}, 2021.

\bibitem{soomro2012ucf101}
Khurram Soomro, Amir~Roshan Zamir, and Mubarak Shah.
\newblock Ucf101: A dataset of 101 human actions classes from videos in the
  wild.
\newblock {\em arXiv preprint arXiv:1212.0402}, 2012.

\bibitem{souvcek2019transnet_shot_det}
Tom{\'a}{\v{s}} Sou{\v{c}}ek, Jaroslav Moravec, and Jakub Loko{\v{c}}.
\newblock Transnet: A deep network for fast detection of common shot
  transitions.
\newblock {\em arXiv preprint arXiv:1906.03363}, 2019.

\bibitem{srivastava2015lstm_unsupervised}
Nitish Srivastava, Elman Mansimov, and Ruslan Salakhudinov.
\newblock Unsupervised learning of video representations using lstms.
\newblock In {\em International conference on machine learning}, pages
  843--852. PMLR, 2015.

\bibitem{tang2020asynchronous_long_vid_pre}
Jiajun Tang, Jin Xia, Xinzhi Mu, Bo Pang, and Cewu Lu.
\newblock Asynchronous interaction aggregation for action detection.
\newblock In {\em European Conference on Computer Vision}, pages 71--87.
  Springer, 2020.

\bibitem{tang2018fast_shot_det}
Shitao Tang, Litong Feng, Zhanghui Kuang, Yimin Chen, and Wei Zhang.
\newblock Fast video shot transition localization with deep structured models.
\newblock In {\em Asian Conference on Computer Vision}, pages 577--592.
  Springer, 2018.

\bibitem{tang2018non_long_vid_pre}
Yongyi Tang, Xing Zhang, Lin Ma, Jingwen Wang, Shaoxiang Chen, and Yu-Gang
  Jiang.
\newblock Non-local netvlad encoding for video classification.
\newblock In {\em Proceedings of the European Conference on Computer Vision
  (ECCV) Workshops}, pages 0--0, 2018.

\bibitem{tao2020selfsupervised_muti_task}
Li Tao, Xueting Wang, and Toshihiko Yamasaki.
\newblock Selfsupervised video representation using pretext-contrastive
  learning.
\newblock {\em arXiv preprint arXiv:2010.15464}, 2, 2020.

\bibitem{tian2019contrastive_endocing}
Yonglong Tian, Dilip Krishnan, and Phillip Isola.
\newblock Contrastive multiview coding.
\newblock 2020.

\bibitem{tran2018r21d}
Du Tran, Heng Wang, Lorenzo Torresani, Jamie Ray, Yann LeCun, and Manohar
  Paluri.
\newblock A closer look at spatiotemporal convolutions for action recognition.
\newblock In {\em CVPR}, pages 6450--6459, 2018.

\bibitem{van2008tsne}
Laurens Van~der Maaten and Geoffrey Hinton.
\newblock Visualizing data using t-sne.
\newblock {\em Journal of machine learning research}, 9(11), 2008.

\bibitem{vapnik1999nature}
Vladimir Vapnik.
\newblock {\em The nature of statistical learning theory}.
\newblock Springer science \& business media, 1999.

\bibitem{vondrick2016anticipating_future_predict}
Carl Vondrick, Hamed Pirsiavash, and Antonio Torralba.
\newblock Anticipating visual representations from unlabeled video.
\newblock In {\em Proceedings of the IEEE conference on computer vision and
  pattern recognition}, pages 98--106, 2016.

\bibitem{wang2019self_motion}
Jiangliu Wang, Jianbo Jiao, Linchao Bao, Shengfeng He, Yunhui Liu, and Wei Liu.
\newblock Self-supervised spatio-temporal representation learning for videos by
  predicting motion and appearance statistics.
\newblock In {\em Proceedings of the IEEE/CVF Conference on Computer Vision and
  Pattern Recognition}, pages 4006--4015, 2019.

\bibitem{wang2020video_peace}
Jiangliu Wang, Jianbo Jiao, and Yun-Hui Liu.
\newblock Self-supervised video representation learning by pace prediction.
\newblock In {\em ECCV}, pages 504--521. Springer, 2020.

\bibitem{wu2019long_lfb}
Chao-Yuan Wu, Christoph Feichtenhofer, Haoqi Fan, Kaiming He, Philipp
  Krahenbuhl, and Ross Girshick.
\newblock Long-term feature banks for detailed video understanding.
\newblock In {\em Proceedings of the IEEE/CVF Conference on Computer Vision and
  Pattern Recognition}, pages 284--293, 2019.

\bibitem{wu2018ins_dis}
Zhirong Wu, Yuanjun Xiong, Stella~X Yu, and Dahua Lin.
\newblock Unsupervised feature learning via non-parametric instance
  discrimination.
\newblock In {\em CVPR}, pages 3733--3742, 2018.

\bibitem{xie2018s3dg}
Saining Xie, Chen Sun, Jonathan Huang, Zhuowen Tu, and Kevin Murphy.
\newblock Rethinking spatiotemporal feature learning: Speed-accuracy trade-offs
  in video classification.
\newblock In {\em ECCV}, pages 305--321, 2018.

\bibitem{xu2019vcop}
Dejing Xu, Jun Xiao, Zhou Zhao, Jian Shao, Di Xie, and Yueting Zhuang.
\newblock Self-supervised spatiotemporal learning via video clip order
  prediction.
\newblock In {\em CVPR}, pages 10334--10343, 2019.

\bibitem{xu2020gtad}
Mengmeng Xu, Chen Zhao, David~S Rojas, Ali Thabet, and Bernard Ghanem.
\newblock G-tad: Sub-graph localization for temporal action detection.
\newblock In {\em Proceedings of the IEEE/CVF Conference on Computer Vision and
  Pattern Recognition}, pages 10156--10165, 2020.

\bibitem{yang2021beyond}
Xitong Yang, Haoqi Fan, Lorenzo Torresani, Larry~S Davis, and Heng Wang.
\newblock Beyond short clips: End-to-end video-level learning with
  collaborative memories.
\newblock In {\em Proceedings of the IEEE/CVF Conference on Computer Vision and
  Pattern Recognition}, pages 7567--7576, 2021.

\bibitem{yao2020seco}
Ting Yao, Yiheng Zhang, Zhaofan Qiu, Yingwei Pan, and Tao Mei.
\newblock Seco: Exploring sequence supervision for unsupervised representation
  learning.
\newblock {\em arXiv preprint arXiv:2008.00975}, 6(7), 2020.

\bibitem{yao2020prp}
Yuan Yao, Chang Liu, Dezhao Luo, Yu Zhou, and Qixiang Ye.
\newblock Video playback rate perception for self-supervised spatio-temporal
  representation learning.
\newblock In {\em CVPR}, pages 6548--6557, 2020.

\bibitem{you2017lars}
Yang You, Igor Gitman, and Boris Ginsburg.
\newblock Large batch training of convolutional networks.
\newblock {\em arXiv preprint arXiv:1708.03888}, 2017.

\bibitem{yue2015beyond_long_vid_pre}
Joe Yue-Hei~Ng, Matthew Hausknecht, Sudheendra Vijayanarasimhan, Oriol Vinyals,
  Rajat Monga, and George Toderici.
\newblock Beyond short snippets: Deep networks for video classification.
\newblock In {\em Proceedings of the IEEE conference on computer vision and
  pattern recognition}, pages 4694--4702, 2015.

\bibitem{zhang2021srtc}
Lin Zhang, Qi She, Zhengyang Shen, and Changhu Wang.
\newblock How incomplete is contrastive learning? an inter-intra variant dual
  representation method for self-supervised video recognition.
\newblock {\em arXiv preprint arXiv:2107.01194}, 2021.

\bibitem{zhang2016colorful}
Richard Zhang, Phillip Isola, and Alexei~A Efros.
\newblock Colorful image colorization.
\newblock In {\em ECCV}, pages 649--666. Springer, 2016.

\bibitem{zhao2019hacs}
Hang Zhao, Zhicheng Yan, Lorenzo Torresani, and Antonio Torralba.
\newblock Hacs: Human action clips and segments dataset for recognition and
  temporal localization.
\newblock {\em arXiv preprint arXiv:1712.09374}, 2019.

\bibitem{zhukov2020learning_t_order}
Dimitri Zhukov, Jean-Baptiste Alayrac, Ivan Laptev, and Josef Sivic.
\newblock Learning actionness via long-range temporal order verification.
\newblock In {\em European Conference on Computer Vision}, pages 470--487.
  Springer, 2020.

\end{thebibliography}
}

\newpage
\onecolumn
\appendix
\renewcommand{\thetable}{A\arabic{table}}
\renewcommand{\thefigure}{A\arabic{figure}}
\renewcommand{\theequation}{A\arabic{equation}}

\section*{Appendix}
In this supplementary material, we first provide detailed theoretical proof for our proposed Gradual Sampling in Section~\ref{sec:proof}. Then the implementation details for pre-training, action recognition, video retrieval, and temporal action localization are introduced in Section~\ref{sec:implement_details}. Finally, we also show more experimental results and visualizations in Section~\ref{sec:additional_results}.

\section{Theoretical Analysis of the Gradual Sampling Strategy}
\label{sec:proof}
In this section, we provide a theoretical understanding of the proposed Gradual Sampling (GS) strategy from the view of generalization analysis, which is commonly used in the literature of learning theory~\cite{vapnik1999nature}. For the sake of simplicity of analysis, we abstract the key points from the GS strategy and make the strategy more math-friendly. As mentioned in the main content of this paper, we divide the training data into two groups, one with small variance (denoted by $\widehat\D_s$) and another one with large variance (denoted by $\widehat\D_l$). At the beginning of training, the sampled clips are considered as examples with small variance since its sampling window size is small according to the definition of $\hat{\Delta}_{\text{max}}(\alpha)$. 
This is reasonable because when the window size is small, the sampled clips are usually similar. While during the later training epochs, the sampled clips could be examples either with large or with small variance since the sampling window size is large and thus it could sample very different clips.
From the viewpoint of optimization, we characterize the difficulty of examples by their variance in gradients. For instance, given two types of examples that are sampled from two different distributions $\D_s$ and $\D_l$, 
it is easy to learn a prediction function from $\D_s$ than from $\D_l$, if 
\begin{align}\label{cond:1:var}
    \E_{\xi\sim\D_s}\left[\|\nabla \ell(w;\xi) - \nabla \F_s(w)\|^2 \right] \leq \E_{\zeta\sim\D_l}\left[\|\nabla \ell(w;\zeta) - \nabla \F_l(w)\|^2 \right],
\end{align}
where $\xi$ and $\zeta$ are the data examples, $w$ is the model parameter, $\ell$ the loss function and 
\begin{align}\label{cond:2:bias}
\nabla\F_s(w) = \E_{\xi\sim\D_s}\left[\nabla\ell(w;\xi)\right], \quad \nabla\F_l(w) = \E_{\zeta\sim\D_l}\left[\nabla\ell(w;\zeta)\right].
\end{align}
In the remaining of this section, we first give the preliminary, then we present the main result in a theorem. All the proofs are included at the end of this section.

\subsection{Preliminary}
To make it easy for our analysis, we formulate the target task as a optimization problem as follows, where the target distribution is a mixture of distributions $\D_s$ and $\D_l$, with a mixture probability $p\in[0,1]$. Formally, the optimization is defined as
\begin{align}\label{pro:opt}
    \min_{w\in\R^d}\Lmath(w) := (1-p)\F_s(w) + p\F_l(w),
\end{align}
where $w$ is the model parameter to be learned, and the loss functions for simple examples and difficult examples are respectively given by
\begin{align}\label{pro:opt:loss}
\F_s(w) = \E_{\xi\sim\D_s}\left[\ell(w;\xi)\right], \quad \F_l(w) = \E_{\zeta\sim\D_l}\left[\ell(w;\zeta)\right].
\end{align}
Here the data examples $\xi$ and $\zeta$ are the data examples that follow  distributions $\D_s$ and $\D_l$ respectively, $\ell$ is a general loss function that can be a single loss or a combined loss of several loss functions. The problem (\ref{pro:opt}) is known as risk minimization (RM). Since the distributions $\D_s$ and $\D_l$ are usually unknown, it is difficult to obtain the loss function $\Lmath(w)$ explicitly. In stead of RM, one can consider its empirical version, which is known an empirical risk minimization (ERM):
\begin{align}\label{pro:opt:erm}
    \min_{w\in\R^d}\widehat\Lmath(w) := (1- p)\widehat\F_s(w) + p\widehat\F_l(w),
\end{align}
where 
\begin{align}\label{pro:opt:loss:erm}
\widehat\F_s(w) = \frac{1}{n}\sum_{\xi_i\in\widehat\D_s} \ell(w;\xi_i), \quad \widehat\F_l(w) = \frac{1}{m}\sum_{\zeta_j\in \widehat\D_l} \ell(w;\zeta_j).
\end{align}
The set of training data $\widehat\D_s := \{\xi_i, i = 1,\dots, n\}$ is sampled from the distribution $\D_s$, and the set of training data $\widehat\D_l := \{\zeta_j, i = 1,\dots, m\}$ is sampled from the distribution $\D_l$. 
To solve the ERM (\ref{pro:opt:erm}), one of simple yet efficient methods is SGD, whose key updating step is given by
\begin{align}
    w_{t+1} = w_t - \eta \nabla_w g(w_t;\xi_{i_t}, \zeta_{i_t}), ~~t = 0,1,\dots, 
\end{align}
where $\eta>0$ is the learning rate, and $\nabla_w g(w;\xi,\zeta)$ the stochastic gradient of $\Lmath(w)$ such that $\E_{\xi,\zeta}[\nabla g(w;\xi,\zeta)] = \nabla \Lmath(w)$. For simplicity, we use $g(w) := g(w;\xi,\zeta)$ in the following analysis. When the variable to be taken a gradient is obvious, we use $\nabla g(w)$ instead of $\nabla_w g(w)$. Similarly, when the randomness is obvious, we use $\E[\cdot]$ instead of $\E_{\xi}[\cdot]$, $\E_{\zeta}[\cdot]$ or $\E_{\xi,\zeta}[\cdot]$.
In this analysis, we are interested in the excess risk bound (ERB), which is a standard measurement of evaluating the solution $\widehat w$ obtained by an algorithm:
\begin{align}
    \Lmath(\widehat w) - \Lmath(w_*),
\end{align}
where $w_*\in\arg\min_{w\in\R^d}\Lmath(w)$ is the optimal solution of problem (\ref{pro:opt}).

For the convenience of analysis, we make the following widely used assumptions for the loss function. 
\begin{ass}[Polyak-\L ojasiewicz condition~\cite{polyak1963gradient}]\label{ass:PL}
There exists a constant $\mu>0$ such that \begin{align*}
2\mu (\Lmath(w) - \Lmath(w_*)) \le \|\nabla \Lmath(w)\|^2, \quad \forall w\in\R^d,
\end{align*}
where $w_* \in\arg\min_{w\in\R^d} \Lmath(w)$ is a optimal solution.
\end{ass}
\begin{ass}[Smoothness~\cite{opac-b1104789}]\label{ass:smooth}
$\Lmath(w)$ is smooth with an $L$-Lipchitz continuous gradient, i.e., it is differentiable and there exists a constant $L>0$ such that 
\begin{align*}
    \|\nabla \Lmath(w)  - \nabla \Lmath(w')\|\leq L\|w - w'\| ,\forall w, w'\in\R^d.
\end{align*}
\end{ass}
Assumption~\ref{ass:smooth} says the objective function $\Lmath(w)$ is smooth with module parameter $L>0$. This assumption has an equivalent expression~\cite{opac-b1104789}: 
    $\Lmath(w) - \Lmath(w') \le \langle \Lmath(w'), w - w' \rangle + \frac{L}{2}\|w-w'\|^2, \quad \forall w, w' \in\R^d$.

We assume that the difference between $\F_s$ and $\F_l$ is captured by $\hat{\Delta}$ from the following assumption.
\begin{ass}\label{ass:f:g}
There exists $\hat{\Delta}\ge0$ such that
\begin{align*}
    \max\limits_{w\in\R^d} \|\nabla \F_s(w) - \nabla\F_l(w)\| \leq \hat{\Delta}.
\end{align*}
\end{ass}
Following the above definition of difficult examples, we assume the following variance structure for stochastic gradients for distribution $\D_s$ and $\D_l$.
\begin{ass}[Bounded variance~\cite{ghadimi2013stochastic}]\label{ass:grad}
The stochastic gradient of $\F_s(w)$ is unbiased and variance bounded. That is, $\E_{\xi\sim\D_s}[\nabla \ell(w;
\xi)] = \nabla \F_s(w)$ and  there exists a constant $\sigma^2>0$, such that
\begin{align*}
\E_{\xi\sim\D_s}\left[\|\nabla \ell(w;
\xi) - \nabla \F_s(w)\|^2 \right] \leq & \sigma^2. 
\end{align*}
\end{ass}
\begin{ass}[Weak Growth Condition~\cite{bertsekas1995neuro, bottou2018optimization}]\label{ass:grad:Fb}
The stochastic gradient of $\Lmath(w)$ is unbiased and variance bounded. That is, $\E_{\xi\sim\D_s}\E_{\zeta\sim\D_l}[\nabla g(w)] = \nabla \Lmath(w)$ and there exists a constant $\sigma^2>0$, such that
\begin{align*}
\E_{\xi\sim\D_l}\E_{\zeta\sim\D_l}\left[\|\nabla g(w) - \nabla \Lmath(w)\|^2 \right] \leq & \frac{ h}{2}\|\nabla \Lmath(w)\|^2 + \sigma^2,
\end{align*}
where $h \gg 1$ is a large constant.
\end{ass}
Assumptions~\ref{ass:grad} and \ref{ass:grad:Fb} imply that $\E_{\zeta\sim\D_l}[\nabla \ell(w;
\zeta)] = \nabla \F_l(w)$ and 
\begin{align}
\E_{\xi\sim\D_l}\left[\|\nabla \ell(w;
\zeta) - \nabla \F_l(w)\|^2 \right] \leq & D + \sigma^2, 
\end{align}
where $D :=\frac{h}{2p^2}\|\nabla \Lmath(w)\|^2 + \frac{(1-p)^2}{p^2}\left(\sigma^2 - E_{\xi\sim\D_s}\left[\|\nabla \ell(w;
\xi) - \nabla \F_s(w)\|^2 \right] \right) > 0$. Please note that $h$ is a very large constant, thus we can consider that the variance for difficult examples is much larger than the variance for simple examples.

 As indicated by the variance structures, stochastic gradients from $\D_l$ exhibit significantly larger variance than those from $\D_s$, particularly at the beginning of the optimization. Hence, it may not be a good idea to run the standard SGD to optimize $\Lmath(w)$. Instead, we could divide the training process into two phases. In the first phase, we will optimize $\Lmath(w)$ using the SGD using the easy examples sampled from distribution $\D_s$. In this way, we could avoid the potentially variance arising from $\D_l$, of course, at the price of bias. In the second phase, when we already received a good solution, we will run the standard SGD to optimize $\Lmath(w)$. Since the solution received from phase I already has excess risk, we will not suffer from the large variance arising from distribution $\D_l$.

\subsection{Theoretical Analysis}
Before the mathematically analysis, we give the following formal version of Theorem {\color{red} 1}, showing that the proposed GS strategy has better generalization than the random sampling (RS) strategy under some mild assumptions. 
\begin{thm}[Formal Version of Theorem {\color{red} 1}]\label{thm:main:formal}
Under Assumptions \ref{ass:PL},  \ref{ass:smooth},  \ref{ass:f:g}, \ref{ass:grad}, \ref{ass:grad:Fb},  we have the following two ERB for RS and GS, respectively. \\ 
(1) for the output of RS $\widehat w_{rs}$, by setting the learning rate $\eta \leq  1/[L(1 + hp)]$, the we have
\begin{align*}
    \E\left[\Lmath(\widehat w_{rs}) - \Lmath(w_*)\right] \leq \exp(-\eta\mu (n+m))(\Lmath(w_0) - \Lmath(w_*)) + \frac{\eta L \sigma^2}{2\mu} \leq O\left( \Lmath(w_0) - \Lmath(w_*)\right).
\end{align*}\\
(2) for the output of GS $\widehat w_{gs}$, by setting the learning rates $\eta_1 = 1/L$ in the first phase and $\eta \leq  1/[L(1 + hp)]$ in the second phase, the we have
\begin{align*}
\E\left[\Lmath(\widehat w_{gs}) - \Lmath(w_*)\right] \leq  O\left(\frac{\sigma^2 L\log(n)}{\mu^2n}+ \frac{p^2\hat{\Delta}^2}{\mu}\right).
\end{align*}
\end{thm}

\subsubsection{Proof of Theorem~\ref{thm:main:formal} (1)}
As the first step, we analyze the RS for optimizing $\Lmath(w)$, where uses both the examples from $\D_s$ and $\D_l$.
\begin{proof}
For the sake of simplicity, let 
denote by $\nabla g(w)$ the stochastic gradient of $\Lmath(w)$ such that $\E[\nabla g(w)] = \nabla \Lmath(w)$. Then the update of SGD for $w_{t+1} = w_t - \eta \nabla g(w_t)$ for $t=0,1,2,\dots$. By the smoothness of function $\Lmath$ from Assumption~\ref{ass:smooth}, we have
\begin{align}
\nonumber&\E[\Lmath(w_{t+1}) - \Lmath(w_t)] \\
\nonumber\leq &\E[ \langle w_{t+1} - w_t, \nabla \Lmath(w_t)\rangle] + \frac{L}{2}\E[\|w_{t+1} - w_t\|^2]\\
\nonumber=& -\eta\E[\langle \nabla g(w_t), \nabla \Lmath(w_t)\rangle] + \frac{\eta^2 L}{2}\E[\|\nabla g(w_t)\|^2] \\
\nonumber= &  - \eta\left(1-\frac{\eta L}{2}\right)\|\nabla \Lmath(w_t)\|^2 + \frac{\eta^2 L}{2}\E[\|\nabla g(w_t)- \nabla \Lmath(w_t)\|^2]\\
\le &  - \eta\left(1-\frac{\eta L}{2}\right)\|\nabla \Lmath(w_t)\|^2 + \frac{\eta^2 L}{2}\E\left[\frac{ h}{2}\|\nabla \Lmath(w_t)\|^2 + \sigma^2\right],
\end{align}
where the last inequality uses Assumptions~\ref{ass:grad}  and~\ref{ass:grad:Fb}. Due to Assumption~\ref{ass:PL},
\begin{align}
\E[\Lmath(w_{t+1}) - \Lmath(w_t)] \le   - \eta\left(1-\frac{\eta L(1+h)}{2}\right)\|\nabla \Lmath(w_t)\|^2 + \frac{\eta^2 L \sigma^2}{2}.
\end{align}
By selecting $\eta \leq \eta_* := 1/[L(1 + h)]$, we have
\begin{align}
\E\left[\Lmath(w_{t+1}) - \Lmath(w_*)\right] \leq \left(1 - \eta\mu\right)\E\left[\Lmath(w_t) - \Lmath(w_*)\right] + \frac{\eta^2 L\sigma^2}{2}.
\end{align}
and therefore
\begin{align}
    \E\left[\Lmath(w_{n+m+1}) - \Lmath(w_*)\right] \leq \exp(-\eta\mu(n+m))(\Lmath(w_0) - \Lmath(w_*)) + \frac{\eta L \sigma^2}{2\mu}.
\end{align}
\end{proof}
{\bf Remark} Since $h$ is large enough, implying that $\eta_*:=1/[L(1 + hp)]$ is small enough, such that
\[
    \exp(-\eta_*\mu (n+m)) \geq \frac{L \sigma^2}{2\mu^2 (n+m)(\Lmath(w_0) - \Lmath(w_*))},
\]
we have
\[
    \E\left[\Lmath(w_{n+m+1}) - \Lmath(w_*)\right] \leq \exp(-\eta_*\mu (n+m))(\Lmath(w_0) - \Lmath(w_*)) + \frac{\eta_* L\sigma^2}{2\mu}.
\]
Consider the special case when $n+m = (hp+1)\kappa$ where $\kappa := L/\mu$ (now $\eta_* = \frac{1}{(n+m)\mu}$), and
\[
    e^{-1} \geq \frac{\sigma^2}{2\mu h\left(\Lmath(w_0) - \Lmath(w_*)\right)}
\]
we have
\[
\E\left[\Lmath(w_{n+m+1}) - \Lmath(w_*)\right] \leq \frac{\Lmath(w_0) - \Lmath(w_*)}{e} + \frac{\sigma^2}{2\mu(h+1)}.
\]
We can see that, due to the large variance arising from $\D_l$, we did not receive a significant reduction in the objective even after $n+m$ iterations when applying RS strategy.


\subsubsection{Proof of Theorem~\ref{thm:main:formal} (2)}
\begin{proof}
In the first phase, we run the optimization using the examples sampled from the distribution $\D_s$.
For the sake of simplicity, let the training examples $\xi_{i_t}, i_t=1,\dots,n'$ are sampled from distribution $\D_s$.  
Then the update of SGD for $w_{t+1} = w_t - \eta_1 \nabla \ell(w_t;\xi_{i_t})$ for $i_t=0,1,2,\dots$. By the smoothness of function $\Lmath$ from Assumption~\ref{ass:smooth}, we have
\begin{align}
\nonumber&\E[\Lmath(w_{t+1}) - \Lmath(w_t)] \\
\nonumber\leq &\E[ \langle w_{t+1} - w_t, \nabla \Lmath(w_t)\rangle] + \frac{L}{2}\E[\|w_{t+1} - w_t\|^2]\\
\nonumber=& -\eta_1\E[\langle \nabla \ell(w_t;\xi_{i_t}), \nabla \Lmath(w_t)\rangle] + \frac{\eta_1^2 L}{2}\E[\|\nabla \ell(w_t;\xi_{i_t})\|^2] \\
\nonumber= & \frac{\eta_1}{2}\|\nabla \F_s(w_t) - \nabla \Lmath(w_t)\|^2  - \frac{\eta_1}{2}\|\nabla \Lmath(w_t)\|^2 - \frac{\eta_1 (1 - \eta_1 L)}{2}\E[\|\nabla \F_s(w_t)\|^2] \\
&+ \frac{\eta_1^2 L}{2}\E[\|\nabla \ell(w_t;\xi_{i_t})- \nabla \F_s(w_t)\|^2]. 
\end{align}
where the last inequality uses $\E[\nabla g(w_t;\xi_t)]=\nabla g(w_t)$. 
Due to Assumptions~\ref{ass:f:g}, \ref{ass:grad}, problem definition~\ref{pro:opt} and $\eta_1 \le 1/L$, we have
\begin{align}
\E[\Lmath(w_{t+1}) - \Lmath(w_t)] 
\leq  \frac{\eta_1 p^2\hat{\Delta}^2}{2} + \frac{\eta_1^2\sigma^2L}{2}  - \frac{\eta_1}{2}\|\nabla \Lmath(w_t)\|^2
\end{align}
Since $\Lmath(\cdot)$ is a $\mu$-PL function under Assumption~\ref{ass:PL}, we have
\begin{align}
    \E[\Lmath(w_{t+1}) - \Lmath(w_t)] \leq -\eta_1\mu\E[\left(\Lmath(w_t) - \Lmath(w_*)\right)] + \frac{\eta_1 p^2\hat{\Delta}^2}{2} + \frac{\eta_1^2\sigma^2L}{2}
\end{align}
and thus
\begin{align}\label{phase:1}
\E[\Lmath(w_{n'+1}) - \Lmath(w_*)] \leq \exp\left(-\eta_1\mu n'\right)\left(\Lmath(w_0) - \Lmath(w_*)\right) + \frac{p^2\hat{\Delta}^2}{2\mu} + \frac{\eta_1\sigma^2L}{2 \mu}.
\end{align}

In the second phase, we analyze the standard SGD for optimizing $\Lmath(w)$ by using the solution of the first phase as the initial solution of SGD. 
The proof is similar to proof of Theorem~\ref{thm:main:formal} (2). By using the result of (\ref{phase:1}), we have
 \begin{align*}
    \E\left[\Lmath(w_{n+m+1}) - \Lmath(w_*)\right] \leq \exp(-\eta \mu n'')\left(\exp\left(-\eta_1\mu n'\right)\left(\Lmath(w_0) - \Lmath(w_*)\right) + \frac{p^2\hat{\Delta}^2}{2\mu} + \frac{\eta_1\sigma^2L}{2 \mu}\right) + \frac{\eta L \sigma^2}{2\mu}.
\end{align*}
When $\hat{\Delta} = 0$ (i.e. the gradients for simple examples and for difficult examples are same), since $\eta = O(1/h)$ is very small, then by letting $\eta_1 = \frac{1}{\mu n'}\log\left( \frac{2\mu^2n'\left(\Lmath(w_0) - \Lmath(w_*)\right)}{\sigma^2 L}\right) \le 1/L$ with $n' = n$, we have
 \begin{align*}
    \E\left[\Lmath(w_{n+m+1}) - \Lmath(w_*)\right] \leq  O\left(\frac{\sigma^2 L\log(n)}{\mu^2n}\right).
\end{align*}
When $\hat{\Delta} \neq 0$ (i.e. the gradients for simple examples and for difficult examples are not same), since $\eta = O(1/h)$ is very small, then by letting $\eta_1 = \min\left(1/L, p^2\hat{\Delta}^2/(2\sigma^2L) \right)$ and $n' \geq \frac{1}{\eta_1\mu} \log\left(\frac{4\mu \left(\Lmath(w_0) - \Lmath(w_*)\right) }{p^2\hat{\Delta}^2}\right)$, we have
 \begin{align*}
    \E\left[\Lmath(w_{n+m+1}) - \Lmath(w_*)\right] \leq  O\left(\frac{p^2\hat{\Delta}^2}{\mu}\right).
\end{align*}

\end{proof}

\section{Implementation Details}
\label{sec:implement_details}
\subsection{Pre-training}
We pre-train all models based on the SimCLR~\cite{chen2020simclr} framework and set $\tau=0.1$.
Three different architectures (S3D-G, R(2+1)D-10, R3D-18) are employed as the encoder $f$. The visual projection head $g$ and topical projection head $h$ each have two hidden layers with 128 output dimensions. In addition, the topical predictor $\phi$ also contains two hidden layers, while the output dimension is 1 for topical consistency prediction. The hyperparameter of Focal Loss in $\mathcal{L}_{\text{TP}}$, \ie, $\gamma$, is set to 0.5.
We adopt standard augmentations used in contrastive approaches for data transformations, \eg, random cropping, color distortion, and horizontal flipping.
We use LARS~\cite{you2017lars} optimizer and set the base learning rate to 0.3. The training learning rate satisfies: $\texttt{LearningRate}=0.3\times \texttt{BatchSize}/256$.
In pre-training, the learning rate is first linearly increased to $\texttt{LearningRate}$ and then decayed with the cosine schedule without restarts.
The batch size is respectively set to 1024, 512, and 1024 for S3D-G, R(2+1)D-10, and R3D-18 networks.
%
For saving the computational costs, each input clip contains 16 frames for S3D-G and R(2+1)D-10, and 8 frames for R3D-18. The spatial resolution is $112\times112$. For ablation experiments, we only train 120 and 50 epochs on HACS and UK400 datasets, respectively.
Our final reported performances are pre-trained for 600 and 500 epochs on these two datasets.

\subsection{Action Recognition}
We fine-tune the models pre-trained by HiCo on UCF101 and HMDB51. If not specified, the input size of the video clips is set to 16$\times$112$\times$112, which is consistent with the pre-training stage. 
For optimizer, we utilize Adam~\cite{kingma2014adam} with batch sizes 1024, 256, 128 for S3D-G~\cite{xie2018s3dg}, R(2+1)D-10~\cite{tran2018r21d}, and R3D18~\cite{hara2018r3d}, respectively. 
The learning rate for these three backbones is set to 0.002, 0.00025, and 0.0002 and decay with a cosine annealing schedule.  
We train 300 epochs on both datasets and adopt the same training strategies for fully fine-tuning and linear fine-tuning. 
In inference, we obtain final predictions by averaging scores from 10 uniformly sampled temporal clips.

\subsection{Video Retrieval}
For nearest-neighbor video retrieval, we first use the pre-trained models without fine-tuning to extract features for both the training set and testing set. Each video will obtain 10 feature vectors by 10 uniformly sampled video clips. Then we average these features for each video and perform \textit{L2} normalization on averaged features. Finally, for each testing video, we calculate its cosine similarities with all training videos. 
The evaluation metric is Recall at $k$ (R@$k$), \ie, a correct retrieval refers to that the top $k$ nearest neighbours contain the correct class.

\subsection{Temporal Action Localization}
Temporal Action Localization(TAL) aims to generate temporal proposals which contain the action instances. 
Two metrics are used to evaluate the generated proposals: AR and AUC. 
The former is Average Recall rate with different tIoU thresholds, and the AUC is calculated by the area under the AR vs. Average Number of proposals (AN) curve, and the AN is varied from $0$ to $100$. 
For each video, we directly adopt the pre-trained models to extract 100 temporally uniform features as the input of BMN~\cite{lin2019bmn}.
The optimizer for BMN is Adamw~\cite{loshchilov2017adamw}, with a learning rate of 0.001 and a weight decay of 1e-6. The learning rate decays with a cosine annealing schedule.
We train BMN for 10 epochs and set the training batch size to 128.
\begin{figure}
\centering
\begin{minipage}[t]{0.48\linewidth}
    \centering
    \includegraphics[width=1.0\linewidth]{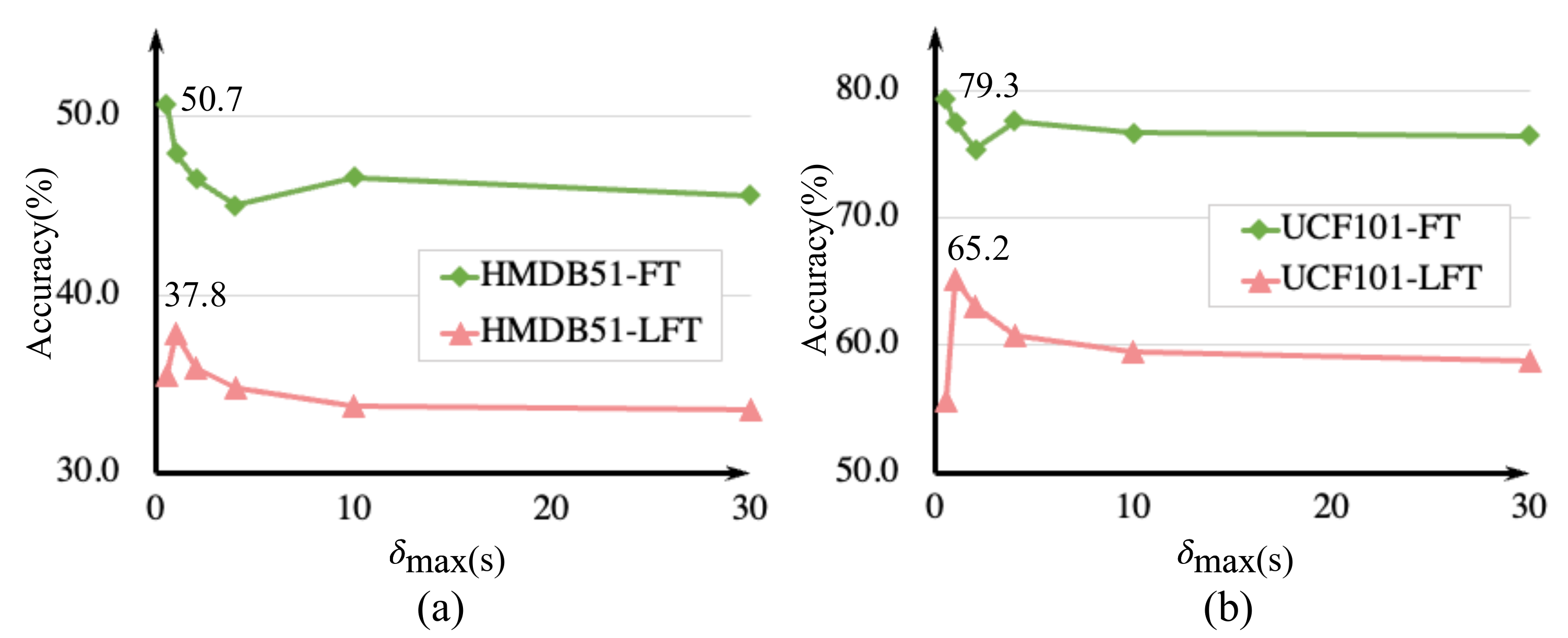}
    \caption{Different $\delta_{\text{max}}$, \ie, the maximum distance between two sampled clips for visual consistency learning. The backbone is S3D-G.}
    \label{fig:dis_vcl}
\end{minipage}
\hspace{0.8em}
\begin{minipage}[t]{0.48\linewidth}
    \centering
    \includegraphics[width=1.0\linewidth]{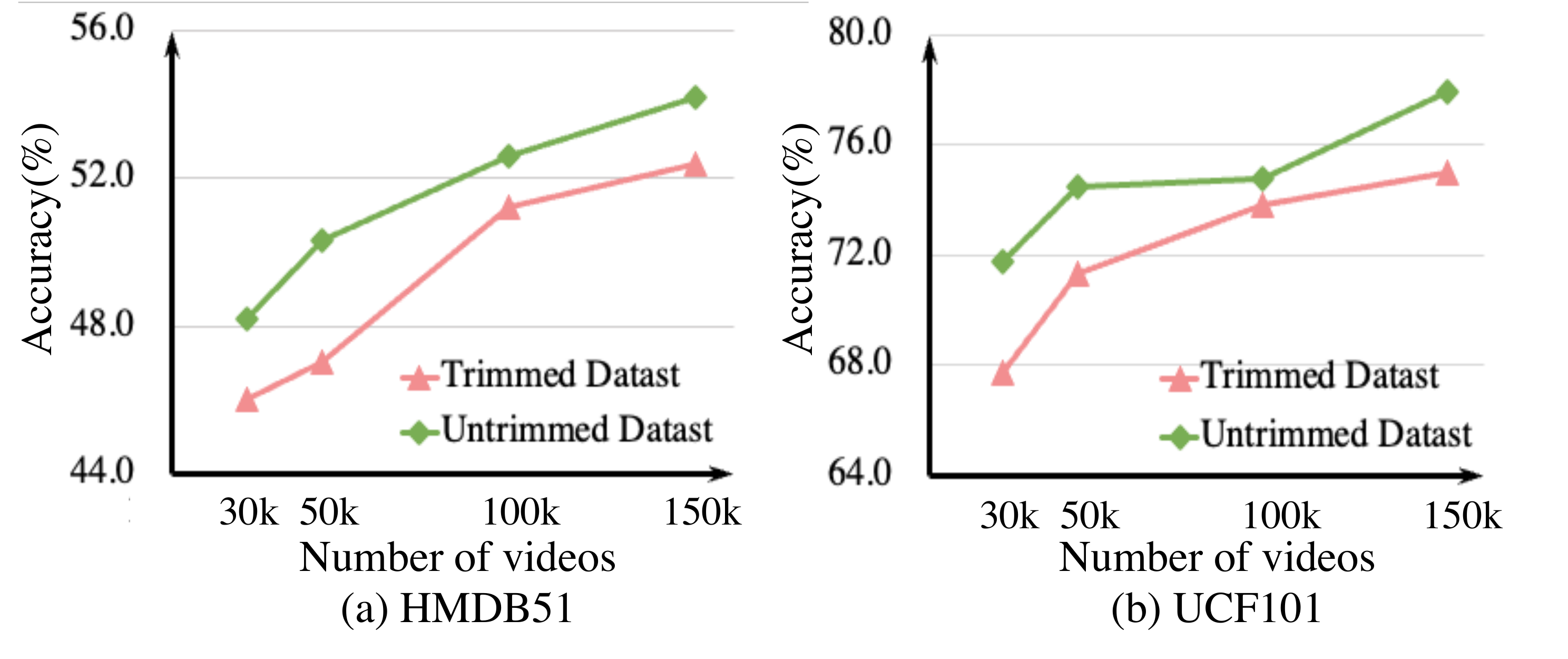}
    \caption{Linear fine-tuning performance comparisons with different numbers of videos for pre-training based on HiCo. S3D-G is employed as the backbone here.}
    \label{fig:vid_num}
\end{minipage}

\end{figure}



\section{Additional Experimental Results}
\label{sec:additional_results}
\subsection{Different temporal distance for VCL}
We propose a simple $\delta_{\text{max}}$ to constrain the maximum distance between two sampled positive clips for visual consistency learning.
To qualify the impact of $\delta_{\text{max}}$, we evaluate different $\delta_{\text{max}}$ based on standard contrastive learning framework, the performance curves on action recognition task are shown in Figure~\ref{fig:dis_vcl} (a) and (b). 
We can observe that increasing $\delta_{\text{max}}$ may hurt both fully fine-tuning and linear fine-tuning accuracy. However, the peak of linear fine-tuning appears at $\delta_{\text{max}}=1s$ on both datasets.
One possible reason is that forcing two semantic unrelated long-range clips to share the same feature embedding will confuse the network. 
Conversely, for two almost identical clips, the network can find shortcuts easily between them and fails to learn powerful representations.

\subsection{Dataset Scales}
We randomly select trimmed videos from the K400 dataset and find their untrimmed versions to generate multiple mini datasets with different scales but the same source. 
The S3D-G is pre-trained with HiCo on these datasets with the same training iteration.
Figure~\ref{fig:vid_num} shows the linear evaluation for the learned representations on both HMDB51 and UCF101.
We observe that HiCo consistently learns more powerful representations from untrimmed videos. 
%
This demonstrates that our HiCo can be generalized to any untrimmed dataset scale.

\begin{figure*}[t]
    \centering
    \includegraphics[width=1.0\linewidth]{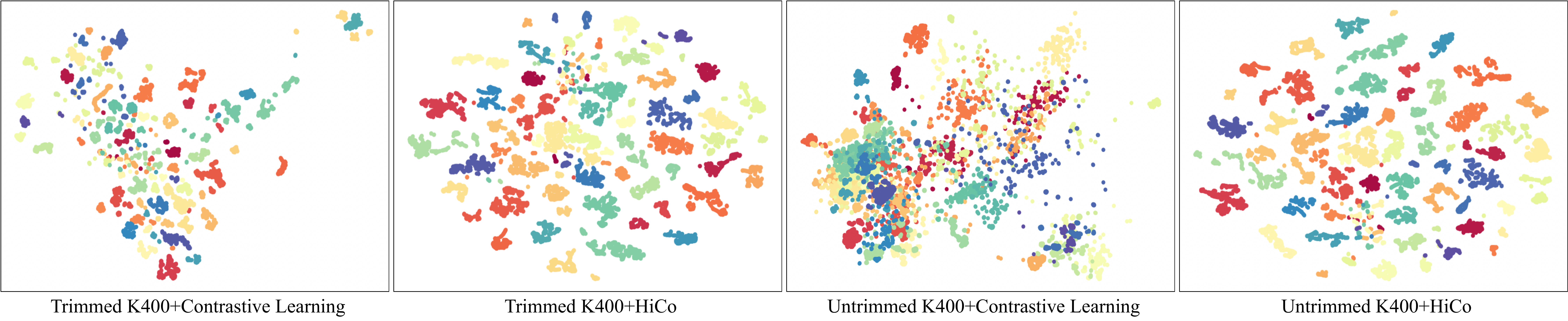}
    \vspace{-3mm}
    \caption{tSNE projection of video features in ActivieyNet-v1.3 dataset. Each color represents an untrimmed video. We present different datasets, \ie, trimmed and untrimmed datasets, with standard contrastive learning and HiCo for pre-training.}
    \vspace{-3mm}
    \label{fig:tsne}
\end{figure*}

\subsection{Visualization}

In Figure~\ref{fig:tsne}, we explore the spatial-temporal representations learned by standard contrastive learning and HiCo on ActivieyNet-v1.3 dataset, using the tool of tSNE projection~\cite{van2008tsne}. The S3D-G network is adopted as a feature extractor, and each color in the figure represents an untrimmed video. 
When pre-training with the standard contrastive learning framework, the separability of the features learned from untrimmed K400 is significantly worse than that from trimmed K400.
This implies forcing different video clips with low visual similarity to share the same feature embedding seriously confuses the network.
In comparison, our HiCo can always learn more robust representations regardless of the trimmed or untrimmed dataset.

\end{document}